%% file: paper.tex
\documentclass[]{bytedance_seed}
\usepackage{tabularx}
\usepackage[toc,page,header]{appendix}

\usepackage{minitoc}
\usepackage{cleveref} 
\usepackage{subcaption}
\usepackage{booktabs}
\usepackage{graphicx}
\usepackage{CJKutf8}
\usepackage{pgfplots}
\usepackage{pgfplotstable}
\usepackage{xcolor}
\usepackage{CJKutf8}
\usetikzlibrary{patterns}
\usepackage{float}
\usepackage{caption}
\usepackage{listings}
\usepackage{xcolor}
\usepackage{tcolorbox}
\tcbuselibrary{listingsutf8}

\lstdefinelanguage{json}{
    basicstyle=\ttfamily\small,
    numbers=none,
    stepnumber=1,
    numbersep=8pt,
    showstringspaces=false,
    breaklines=true,
    frame=single,
    backgroundcolor=\color{gray!5},
    literate=
     *{0}{{{\color{black}0}}}{1}
      {1}{{{\color{black}1}}}{1}
      {2}{{{\color{black}2}}}{1}
      {3}{{{\color{black}3}}}{1}
      {4}{{{\color{black}4}}}{1}
      {5}{{{\color{black}5}}}{1}
      {6}{{{\color{black}6}}}{1}
      {7}{{{\color{black}7}}}{1}
      {8}{{{\color{black}8}}}{1}
      {9}{{{\color{black}9}}}{1}
      {:}{{{\color{black}:}}}{1}
      {,}{{{\color{black},}}}{1}
      {"}{{{\color{black}"}}}{1}
}
\newtcolorbox{promptbox}[1][]{
  colbacktitle=black!60,
  coltitle=white,
  fontupper=\footnotesize,
  colback=gray!5,
  boxsep=5pt,
  left=5pt,
  right=5pt,
  top=5pt,
  bottom=5pt,
  boxrule=1pt,
  title={#1},
  width=0.95\textwidth, % 控制宽度
  enhanced,
  breakable,            % 支持自动分页和换行
}

\input{macro}

\title{How RL Unlocks the Aha Moment in Geometric Interleaved Reasoning}

\author[1,*]{Xiangxiang Zhang}
\author[2,4,*]{Caijun Jia}
\author[1,3,*]{Siyuan Li}
\author[1,*]{Dingyu He}
\author[1,*]{Xiya Xiong}
\author[2,4]{\\ Zheng Sun}
\author[2,4]{Honghao He}
\author[1]{Yuchen Wu}
\author[2,4]{Bihui Yu}
\author[2,4]{Linzhuang Sun}
\author[3,\dag]{Cheng Tan}
\author[2,4,5,\dag]{\\ Jingxuan Wei}

\affiliation{$^{1}$ByteDance, $^{2}$Shenyang Institute of Computing Technology, Chinese Academy of Sciences}
\affiliation{$^{3}$ Westlake University, $^{4}$ University of Chinese Academy of Science, $^{5}$ Key Laboratory of Computing Power Network and Information Security, Ministry of Education, Shandong  Computer Science Center (National Supercomputer Center in Jinan),  Qilu University of Technology  (Shandong Academy of Sciences)}

\contribution[*]{Equal contribution}
\contribution[\dag]{Corresponding author}

\abstract{

\input{sections/000abstract}
}

\newtcolorbox{datasetpromptbox}[1][]{
  enhanced,
  top=0.3em,bottom=0.3em,left=0.5em,right=0.5em,
  toptitle=0.3em,bottomtitle=0.2em,boxsep=0pt,
  colframe=seedblue,     % 边框 = 模板蓝色
  colback=seedblue!10,   % 背景 = 极淡蓝色（好看不刺眼）
  boxrule=0.5pt,
  width=\linewidth,
  title={\footnotesize #1}
}
    
\begin{document}

\maketitle

%不需要目录就注释掉 注意目录不要和第一页放在一块 要有\newpage
%\newpage
%\tableofcontents
%\newpage

\input{sections/010intro}
\input{sections/020related}
\input{sections/030data}
\input{sections/040verify}
\input{sections/050rl}
\input{sections/100conclusion}

\clearpage

\bibliographystyle{unsrt}
\bibliography{main}

\clearpage
\beginappendix
\section{Evaluation Metrics and Protocols}
\label{app:metrics}

\subsection{Evaluation metrics}
\label{app:metrics_overview}

We evaluate multimodal geometry solving along two axes: \emph{solution rigor} (answer correctness) and \emph{diagram quality} (whether the generated GeoGebra construction is usable, aligned with the intended reasoning, and geometrically valid).
Since geometric constructions admit many equivalent realizations, we report both intention-oriented verification scores and surface-level similarity scores.
All automatic judges follow fixed prompts and deterministic execution rules for reproducibility; detailed protocols appear in Appendix~\ref{app:metrics}.

\noindent\textbf{Answer accuracy.}
We extract the final answer $\hat{a}$ from the model output and compare it with the reference answer $a^\star$ under mathematical equivalence.
Accuracy is binary: $1$ if correct (or equivalent) and $0$ otherwise.

\noindent\textbf{Drawing score.}
To assess drawing capability beyond appearance matching, we use a tri-perspective evaluation that targets three complementary failure modes in executable geometry: perceptually unusable renders, intent-shifted constructions, and formally invalid relations.
For each sample, we compute three normalized scores in $[0,1]$ and report their mean as the final drawing score:
\begin{equation}
S_{\text{draw}} = \frac{1}{3}\Big(s_{\text{vis}} + s_{\text{sem}} + s_{\text{form}}\Big).
\label{eq:draw_score}
\end{equation}
Here $s_{\text{vis}}$ comes from the \textbf{Visual Judge} (VLM-Judge), which evaluates perceptual usability by judging the rendered image $I_{\text{render}}$ against the problem diagram $I_{\text{prob}}$ with the solution trace as context.
$s_{\text{sem}}$ comes from the \textbf{Semantic Judge} (Parser-Judge), which parses the program into an intermediate representation $T_{\text{IR}}$ and checks consistency between the reasoning trace and $T_{\text{IR}}$.
$s_{\text{form}}$ comes from the \textbf{Formal Judge} (Assert-Judge), which executes synthesized geometric assertions in the GeoGebra kernel and scores the assertion pass rate.
We also report the three sub-scores separately to diagnose failure patterns such as ``looks right but violates constraints'' or ``formally correct but unreadable''.

\noindent\textbf{Surface similarity metrics.}
For completeness, we additionally report objective similarity scores that quantify surface-level proximity to references.
\emph{Code similarity} measures how close the generated GeoGebra script is to a reference script using BLEU, ROUGE-L, chrF, Edit Distance, and RUBY (higher is better except Edit Distance).
\emph{Image similarity} compares $I_{\text{render}}$ with the reference diagram using PSNR, SSIM, and LPIPS (higher is better except LPIPS).
These similarity metrics are not reliable proxies for geometric correctness under an open-ended solution space, but they help characterize formatting and appearance alignment.

\subsection{Tri-perspective drawing evaluation}
\label{app:tri_eval}

\paragraph{Visual Judge (VLM-Judge).}
We execute the generated GeoGebra program to obtain a rendered image $I_{\text{render}}$.
A vision-language judge receives $(I_{\text{prob}}, I_{\text{render}}, T_{\text{infer}})$ and outputs a normalized score $s_{\text{vis}}\in[0,1]$.
The rubric checks (i) element completeness within the viewport, (ii) connectivity of auxiliary constructions, (iii) legibility, and (iv) structural consistency with the intended diagram.
Rendering failure yields $s_{\text{vis}}=0$.

\paragraph{Semantic Judge (Parser-Judge).}
We parse the generated program into a structured intermediate representation $T_{\text{IR}}$ that lists primitives and relations together with attributes and dependencies.
A judge checks whether the reasoning trace $T_{\text{infer}}$ is consistent with $T_{\text{IR}}$ and outputs $s_{\text{sem}}\in[0,1]$.
Parsing failure or missing required entities yields $s_{\text{sem}}=0$.

\paragraph{Formal Judge (Assert-Judge).}
We synthesize a set of executable assertions from the problem specification and evaluate them in the GeoGebra kernel on the constructed state.
We define the formal score as a weighted pass rate:
\begin{equation}
s_{\text{form}} = \frac{\sum_{j} w_j \cdot \mathbb{1}[\text{assertion}_j\ \text{passes}]}{\sum_{j} w_j},
\label{eq:assert_score}
\end{equation}
where $w_j$ assigns higher weight to key assertions directly implied by the problem.
Optionally, key-assertion failure triggers $s_{\text{form}}=0$ to prevent partial-credit artifacts.

\subsection{Code similarity metrics}
\label{app:code_sim}

We treat GeoGebra scripts as structured text and compute surface similarity against a reference script.
We tokenize by commands, identifiers, numbers, and delimiters.
We report BLEU, ROUGE-L, chrF, Edit Distance, and RUBY.
RUBY applies lightweight normalization (e.g., whitespace removal and numeric canonicalization) before matching and weights key construction commands more heavily.

\subsection{Image similarity metrics}
\label{app:image_sim}

We compare $I_{\text{render}}$ with a reference diagram using PSNR, SSIM, and LPIPS.
These metrics quantify appearance alignment in pixel space or feature space and are sensitive to non-essential factors such as cropping, global scaling, and label placement.
We therefore use them as diagnostic signals rather than primary measures of geometric correctness.

\section{Comparison with Existing Benchmarks}
\label{subsec:benchmark_comparison}

Table~\ref{tab:benchmark_comparison} positions \textbf{Faire-Bench} among representative multimodal math benchmarks along two dimensions that matter for constructive geometry reasoning.
First, \emph{capability coverage} asks whether a benchmark requires (i) solving the math problem, (ii) producing a diagram as an output, and (iii) solving in an interleaved text–image manner.
Second, \emph{supervision modality} asks whether instances provide aligned \emph{text}, \emph{image}, and \emph{code} signals.
We report each attribute as a binary indicator for clarity.

\noindent\textbf{From answer selection to constructive evidence.}
Most multimodal math benchmarks focus on problem solving with static visual inputs and evaluate models primarily by answer correctness.
Even when generation is considered, the output is rarely required to be a \emph{constructive artifact} that can be executed and checked against geometric constraints.
Faire-Bench instead evaluates a model’s ability to act: it must produce both a correct solution and an executable construction that reconstructs the intended geometric state.

\noindent\textbf{Tri-modal supervision for verifiability.}
A central limitation in prior benchmarks is the lack of aligned program supervision.
Without code, it is difficult to verify whether the reasoning and the final diagram are mutually consistent, and failures often collapse into unscored hallucinations.
Faire-Bench provides aligned \emph{text}, \emph{image}, and \emph{code} for every instance, enabling deterministic executability checks and geometry-aware verification beyond surface similarity.

\noindent\textbf{Interleaved solving with executable grounding.}
Benchmarks that include drawing typically emphasize visualization quality, but do not couple drawing with the full problem-solving process, and thus do not support evaluating interleaved reasoning and construction as a single coherent behavior.
Faire-Bench explicitly supports interleaved text–image solving and grounds that behavior in executable code, which makes intermediate constructions auditable rather than merely narrated.

\noindent\textbf{Takeaway.}
Together, these axes make Faire-Bench a constructive and verifiable benchmark that unifies reasoning, drawing, and interleaved solving under tri-modal supervision, complementing understanding-oriented math datasets and filling the gap between visual reasoning and executable geometry.

\begin{table}[t]
% \small
\centering
% \vspace{-1mm}
\caption{\textbf{Comparison with existing multimodal math benchmarks.}
We contrast capability coverage (solving, drawing, interleaved solving) and supervision modality (text, image, code).}
% \vspace{-2mm}
\label{tab:benchmark_comparison}
\setlength{\tabcolsep}{4mm}
\begin{adjustbox}{max width=0.95\textwidth}
\begin{tabular}{lccccccc}
\toprule
Benchmark &
Solving &
Drawing &
Interleaved &
Text &
Image &
Code \\
\midrule
MathVista~\cite{mathvista}        & \cmark & \xmark & \xmark & \cmark & \cmark & \xmark \\
MathVerse~\cite{mathverse}        & \cmark & \xmark & \xmark & \cmark & \cmark & \xmark \\
MM-MATH~\cite{sun2024mm}           & \cmark & \xmark & \xmark & \cmark & \cmark & \xmark \\
MathScape~\cite{mathscape}        & \cmark & \xmark & \xmark & \cmark & \cmark & \xmark \\
GeoEval~\cite{zhang2024geoeval}          & \cmark & \xmark & \xmark & \cmark & \cmark & \xmark \\
WE-MATH~\cite{wemath}          & \cmark & \xmark & \xmark & \cmark & \cmark & \xmark \\
MMSciBench~\cite{ye2025mmscibench}       & \cmark & \xmark & \xmark & \cmark & \cmark & \xmark \\
MATH2VISUAL~\cite{wang2025generating}      & \cmark & \cmark & \xmark & \cmark & \cmark & \xmark \\
PolyMath~\cite{polymath}         & \cmark & \xmark & \xmark & \cmark & \cmark & \xmark \\
SOLIDGEO~\cite{solidgeo}         & \cmark & \xmark & \xmark & \cmark & \cmark & \xmark \\
GGBench~\cite{wei2025ggbench}          & \xmark & \cmark & \xmark & \cmark & \cmark & \cmark \\
\textbf{Faire-Bench (ours)} & \cmark & \cmark & \cmark & \cmark & \cmark & \cmark \\
\bottomrule
\end{tabular}
% \vspace{-3mm}
\end{adjustbox}
\end{table}

\section{Verifier Prompt Specifications}
\begin{figure*}[t]
\centering
\begin{datasetpromptbox}[VLM-as-Verifier Prompt for Diagram Fidelity]

\footnotesize
You are a GeoGebra diagram quality auditor with strong mathematical expertise.
Your task is to verify whether a rendered diagram is \emph{mathematically correct}, \emph{logically consistent}, and \emph{visually usable}, by jointly examining the problem image, the solution text (including reasoning flow), and the rendered diagram produced from GeoGebra code.

\textbf{Inputs}

You are given:
(1) the original problem image as reference;  
(2) the solution text, including step-by-step reasoning and embedded \texttt{geogebra} code blocks;  
(3) the rendered diagram generated from the code (\textbf{this is the primary object to be evaluated}).

\textbf{Verification Checklist}

Perform the following checks and decide whether the rendered diagram is usable.

\textbf{(1) Contextual intent alignment}  
First identify the drawing intent from the surrounding solution text.
If the text explicitly limits scope (e.g., ``we illustrate case 1 only'' or ``we draw the translated figure''), do \emph{not} penalize missing elements outside this declared scope.
The diagram is considered aligned as long as it faithfully reflects what the current step claims to construct.

\textbf{(2) Mathematical correctness and geometric completeness}  
Verify that all required points, lines, curves, or function plots needed for the reasoning step are present.
Check that defined mathematical relations (e.g., incidence, collinearity, relative position, shape properties) are correctly instantiated.
The geometric relations expressed by the diagram must be consistent with the solution text and the problem specification.

\textbf{(3) Visual connectivity and structural integrity}  
Inspect all auxiliary constructions such as dashed lines, perpendiculars, or projection guides.
Auxiliary elements must be physically connected to their target objects (axes, curves, points).
If any auxiliary line is visibly disconnected or floating, mark the diagram as invalid.

\textbf{(4) Visual clarity and readability}  
Assess whether the diagram is clear and interpretable.
Penalize severe clutter, overlapping elements, blurred rendering, or occluded labels.
Key points and annotations should be legible and unambiguous.

\textbf{(5) Allowed visual deviations}  
Do not penalize distortions caused by automatic axis scaling in GeoGebra.
As long as topological relations are preserved (e.g., an ellipse remains a closed curve), differences in apparent proportions are acceptable.

\textbf{Output Format}

Return a binary decision:
\texttt{1} if the diagram is usable; \texttt{0} otherwise.

If returning \texttt{0}, provide a brief justification in the form:
\texttt{[Error Type] Specific description},  

for example: \texttt{[Mathematical Error] Point (3,2) is shown on line $y=x-2$ but does not satisfy the equation}, or  
\texttt{[Visual Error] The dashed projection from point A does not connect to the x-axis}.

\end{datasetpromptbox}
\caption{Prompt used for the \emph{VLM-as-Verifier} in Faire. The verifier performs pixel-level, mathematical, and logical consistency checks to determine whether a rendered diagram faithfully supports the intended reasoning step.}
\label{fig:vlm_verifier_prompt}
\end{figure*}

\subsection{VLM-as-Verifier.}
Figure~\ref{fig:vlm_verifier_prompt} shows the prompt used for the vision-language verifier in Faire.
Unlike generic preference-based judges, this verifier is explicitly instructed to assess diagram usability through a structured checklist that combines contextual intent, mathematical validity, geometric connectivity, and visual readability.
Crucially, the verifier is designed to tolerate benign visual distortions introduced by automatic coordinate scaling, while remaining strict about topological correctness and logical consistency.
This design enables pixel-level yet mathematically grounded supervision, forming a key component of the post-generation verification pipeline.

\begin{figure*}[ht!]
\centering
\begin{datasetpromptbox}[Parser-Based Semantic Verification Prompt]

\footnotesize
You are an expert verifier specializing in analytic geometry and step-by-step mathematical reasoning.
Your role is to act as a \textbf{semantic geometry validator} that audits whether a constructed diagram faithfully instantiates the reasoning process described in the solution.

\textbf{Task Description}

You are given:
(1) a problem image;
(2) a detailed solution text containing one or more \texttt{ggb\_parser} blocks.

Each \texttt{ggb\_parser} block is a \emph{textual description} of a geometric construction, listing points, lines, curves, intersections, attributes (coordinates, domains, visibility), and relations.
It is \textbf{not executable code}.
Your task is to verify whether this description is semantically consistent with both the problem statement and the mathematical reasoning steps.

\textbf{Verification Checklist}

For \emph{each} \texttt{ggb\_parser} block, perform the following checks.

\emph{(I) Element Accuracy}

\textbf{Points.}
Verify that all point coordinates match those derived in the solution.
List where each point is expected to lie (on which curve, line, or segment), then check whether the parser description is consistent.
Missing or misplaced points constitute an error.

\textbf{Curves.}
Check whether curve definitions (equations, branches, domains) exactly match the derivation.
For example, if the solution specifies a restricted branch of a conic, the parser must reflect the same restriction.

\textbf{Lines.}
Verify that line equations (slope, intercept, tangency) match the computed results.

\textbf{Hidden Elements.}
For invisible objects, ensure that required connections are still materialized.
Hidden carrier lines must still induce the necessary visible segments to avoid disconnected points.

\textbf{Angles.}
If angle elements appear in the parser, verify that they are explicitly required by the problem or solution.
Any unnecessary non-right angle is considered incorrect.

\textbf{Length Relations.}
Exact numeric lengths need not match, but relative ordering must.
For example, if $AC > AB > BC$ in the solution, the construction must preserve this inequality.

\textbf{Polygons.}
Verify polygon naming consistency using coordinate-based cyclic ordering.
Determine the true geometric vertex cycle via centroid and polar-angle sorting, and check whether the given name matches up to rotation or reversal.
Incorrect adjacency implies an invalid polygon.

\emph{(II) Relational Consistency}

Check that all incidence relations (point-on-line, point-on-curve) and proven geometric relations (collinearity, tangency, concurrency) are explicitly reflected in the parser description.

\emph{(III) Internal Consistency}

Extract the diagram description sentence (e.g., ``the figure below shows...'').
Verify that every described element appears in the parser and that no contradictions exist.

\emph{(IV) Completeness}

Based on the original problem image and statement, list all required geometric elements.
If any required element is missing across all parser blocks, the construction is invalid.

\textbf{Output Format}

For each parser block, produce a structured \emph{Diagram Verification Report} including:
(1) element-level analysis;
(2) relation-level consistency;
(3) internal consistency;
(4) completeness;
(5) a list of detected issues (or ``no issues found'').

Conclude with three binary judgments:
(a) corresponds to the solution process;
(b) satisfies the problem intent;
(c) is geometrically correct.

\textbf{Final Decision}: output \texttt{1} if and only if all three judgments are positive; otherwise output \texttt{0}.
If multiple parser blocks exist, the final score is \texttt{1} only if all blocks pass.

\end{datasetpromptbox}
\caption{Prompt used for parser-based semantic verification, which evaluates whether a textual geometric description faithfully instantiates the intended reasoning and relations.}
\label{fig:parser_verifier_prompt}
\end{figure*}

\subsection{Parser-as-Verifier.}

Figure~\ref{fig:parser_verifier_prompt} presents the prompt used for the parser-based verifier in Faire.
This verifier targets semantic consistency between the symbolic geometric description and the reasoning process, operating at a level orthogonal to pixel-based inspection.

Unlike the VLM-based verifier, which evaluates rendered diagrams, the parser-based verifier audits a structured textual description of the construction (\texttt{ggb\_parser}) that enumerates geometric entities, relations, and attributes.
The prompt enforces a fine-grained, checklist-driven comparison between the parsed geometric specification and the mathematical reasoning steps, covering element existence, relational correctness, hidden-structure integrity, and internal consistency.

By grounding verification in explicit coordinates, equations, and relational constraints, the parser-based verifier detects failures that are visually plausible yet semantically incorrect.
It therefore instantiates the semantic alignment signal in our tri-perspective verification framework, complementing visual usability checks and formal geometric assertions.

\begin{figure*}[t]
\centering
\begin{datasetpromptbox}[GGB-as-Verifier Prompt for Formal Geometric Validity]

\footnotesize
You are an expert in GeoGebra and formal geometric verification.
Your task is to analyze a rendered geometric diagram together with its corresponding GeoGebra construction, and to generate executable verification commands that test whether the construction satisfies the geometric relations implied by the diagram.

\textbf{Inputs}

You are given:
(1) an image of the rendered geometric construction;  
(2) a set of GeoGebra commands (\texttt{user\_construction}) that produced the diagram.

\textbf{Task Description}

Your goal is to identify key geometric properties visible in the diagram and translate them into formal GeoGebra boolean expressions.
These expressions will be executed by the GeoGebra kernel to verify whether the construction is mathematically correct.

\textbf{Verification Procedure}

\textbf{(1) Visual relation analysis}  
Inspect the diagram and identify essential geometric relations, such as:
points lying on lines or curves, intersections, collinearity, concurrency, parallelism, perpendicularity, tangency, concyclicity, or region inclusion.

\textbf{(2) Assertion synthesis}  
For each identified relation, generate a corresponding GeoGebra boolean predicate that evaluates to \texttt{true} if the relation holds.
Each predicate must be directly executable on the provided construction.

\textbf{(3) Allowed primitives}  
You may use the following GeoGebra verification functions:
\begin{itemize}
  \item \texttt{AreCollinear(<Point>, <Point>, <Point>)}
  \item \texttt{AreConcurrent(<Line>, <Line>, <Line>)}
  \item \texttt{AreConcyclic(<Point>, <Point>, <Point>, <Point>)}
  \item \texttt{AreParallel(<Line>, <Line>)}
  \item \texttt{ArePerpendicular(<Line>, <Line>)}
  \item \texttt{IsTangent(<Line>, <Conic>)}
  \item \texttt{IsInRegion(<Point>, <Region>)}
\end{itemize}

\textbf{Output Format}

Return a JSON object with a single field:
\begin{itemize}
  \item \texttt{"verification\_code"}: a list of GeoGebra boolean expressions, where each expression verifies one geometric property.
\end{itemize}

All generated expressions must be valid GeoGebra commands and should collectively cover the essential geometric relations depicted in the diagram.

\end{datasetpromptbox}
\caption{Prompt used for the \emph{GGB-as-Verifier} in Faire. The verifier synthesizes executable GeoGebra assertions from visual relations and evaluates them deterministically to enforce formal geometric correctness.}
\label{fig:ggb_verifier_prompt}
\end{figure*}

\subsection{GGB-as-Verifier.}

Figure~\ref{fig:ggb_verifier_prompt} shows the prompt used for the GeoGebra-based verifier in Faire.
This verifier provides a deterministic correctness signal by checking whether the generated construction satisfies formal geometric properties implied by the diagram.

Unlike the visual and parser-based verifiers, which rely on perceptual or semantic judgments, the GGB-as-Verifier operates by synthesizing explicit boolean assertions from the observed geometric relations.
Given the rendered diagram and the corresponding GeoGebra commands, the verifier translates visual relations (e.g., collinearity, perpendicularity, tangency, or region membership) into executable GeoGebra predicates.

These predicates are then evaluated directly by the GeoGebra kernel, yielding a binary outcome that reflects mathematical truth rather than appearance.
As such, the GGB-as-Verifier instantiates the \emph{formal validity} signal in our tri-perspective verification framework, preventing self-consistent but geometrically incorrect constructions from receiving reward.

\begin{table*}[t]
\centering
\small
% \vspace{-1mm}
\caption{Results on GGBench.
Stage-wise interleaved evaluation (higher is better) and image similarity (lower LPIPS is better).}
\vspace{-2mm}
\label{tab:ggbench_results}
\setlength{\tabcolsep}{4.2pt}
\renewcommand{\arraystretch}{1.06}
\resizebox{\linewidth}{!}{
\begin{tabular}{lccccccc}
\toprule
\multirow{2}{*}{Model} &
\multicolumn{3}{c}{Stage-wise scores} &
\multicolumn{3}{c}{Image similarity} &
\multirow{2}{*}{Overall} \\
\cmidrule(lr){2-4}\cmidrule(lr){5-7}
& Planning (VLM-T)$\uparrow$ & Mid (VLM-I-Mid)$\uparrow$ & Final (VLM-I-Res)$\uparrow$
& LPIPS$\times 10^{-2}\downarrow$ & PSNR$\uparrow$ & SSIM$\times 10^{-2}\uparrow$
& VLM-I$\uparrow$ \\
\midrule
\multicolumn{8}{c}{\textit{End-to-end UMMs}} \\
\midrule
Qwen-Image~\cite{wu2025qwen}        & \textemdash & \textemdash & 22.75 & 56.39 & 58.23 & 48.06 & 22.75 \\
Janus~\cite{janus}            & 33.85 & 21.69 & 19.76 & 57.74 & 57.76 & 60.97 & 20.73 \\
Nano Banana~\cite{google2025gemini25}      & 58.54 & 44.83 & 22.81 & 51.85 & 64.53 & 59.51 & 33.82 \\
GPT-image-1~\cite{hurst2024gpt}      & \textemdash & \textemdash & 39.11 & 50.67 & 16.95 & 64.37 & 39.11 \\
\midrule
\multicolumn{8}{c}{\textit{LLMs/LRMs}} \\
\midrule
GPT-4o~\cite{hurst2024gpt}           & 59.73 & 26.19 &  2.66 & 95.43 &  5.45 &  5.69 & 14.43 \\
GLM-4.5V~\cite{hong2025glm}         & 53.32 & 25.63 &  5.02 & 52.91 & 12.19 & 12.94 & 15.33 \\
Qwen3-14B~\cite{yang2025qwen3}        & 58.65 & 39.30 & 12.97 & 78.81 & 23.92 & 24.81 & 26.13 \\
Gemini 2.5 Pro~\cite{google2025gemini25}   & 38.50 & 37.41 & 15.80 & 68.39 & 37.17 & 39.73 & 26.61 \\
DeepSeek-R1~\cite{liu2024deepseek}      & 61.16 & 62.42 & 20.48 & 66.06 & 37.94 & 37.59 & 41.45 \\
GPT-4~\cite{achiam2023gpt}            & 55.66 & 50.39 & 20.30 & 67.35 & 35.26 & 38.31 & 33.04 \\
Qwen3-VL~\cite{yang2025qwen3}         & 56.40 & 49.55 & 23.94 & 39.40 & 52.33 & 58.71 & 36.74 \\
DeepSeek-V3.1~\cite{liu2024deepseek}    & 60.24 & 73.13 & 26.41 & 57.21 & 48.33 & 50.12 & 49.77 \\
Claude Sonnet 4.5~\cite{Anthropic2025ClaudeSonnet4_5} & 61.19 & 77.92 & 30.29 & 52.22 & 51.74 & 50.52 & 54.11 \\
GPT-5~\cite{OpenAI2025GPT5SystemCard}            & \textbf{62.01} & 76.79 & \textbf{37.36} & 49.65 & \textbf{54.80} & 59.49 & 57.08 \\
\midrule
\textbf{Faire} & 60.13 & \textbf{89.11} & 37.14 & \textbf{33.31} & 20.36 & \textbf{75.57} & \textbf{63.13} \\
\bottomrule
\end{tabular}
}
\vspace{-2mm}
\end{table*}

% \section{Category-wise Analysis}
\section{Where Functional Alignment Matters Most}
\label{sec:category_analysis}

\begin{figure*}[t]
  \centering
  \includegraphics[width=\linewidth]{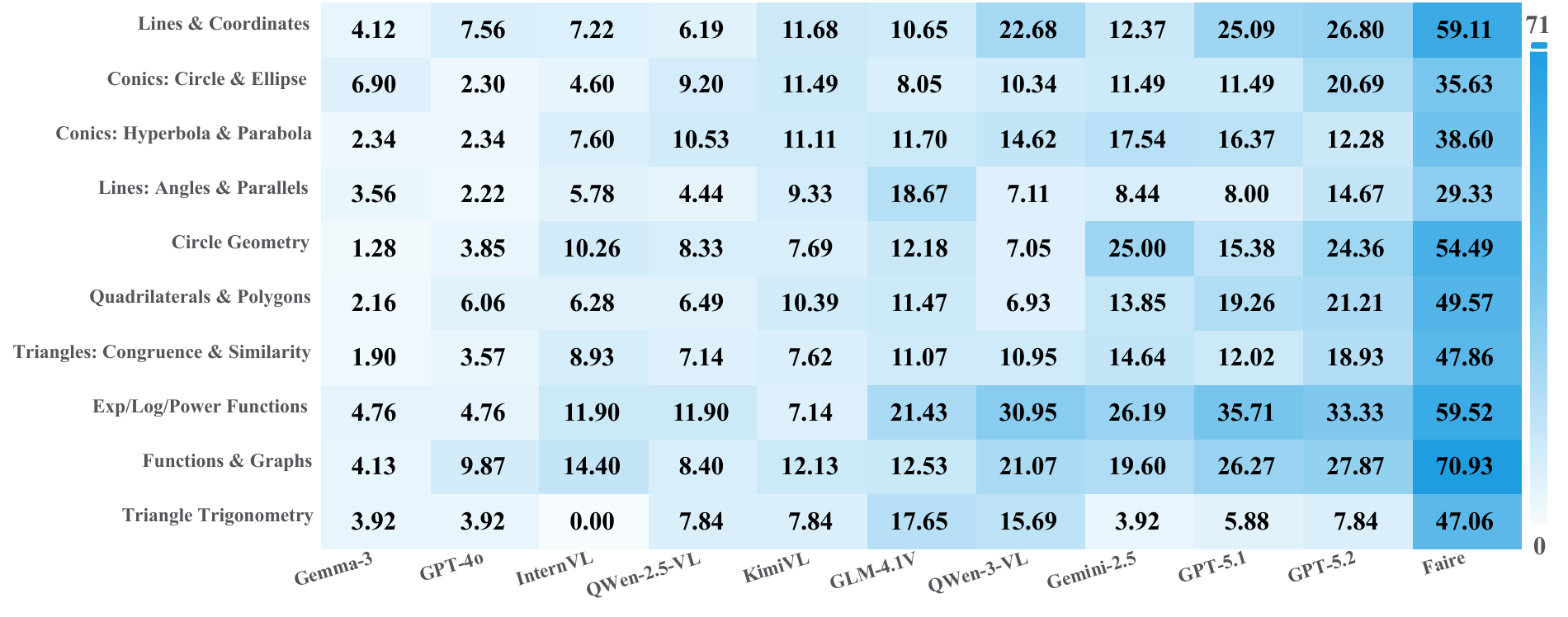}
  \caption{Category-wise verification scores.}
  \vspace{-2mm}
  \label{fig:category_heatmap}
\end{figure*}

Figure~\ref{fig:category_heatmap} reveals a clear visual dichotomy between \emph{systematic failure} and \emph{robust grounding}.
Across geometry-heavy categories, several strong generalist MLLMs form large pale regions in the heatmap, indicating not sporadic errors but a structural inability to ground deductions in constructed geometric states.
For instance, on \emph{Lines \& Coordinates}, Gemini-2.5~\cite{google2025gemini25} reaches only 12.37 and GPT-4o~\cite{hurst2024gpt} drops to 7.56, whereas Faire achieves 59.11, representing nearly a five-fold improvement.
A similar pattern appears on \emph{Conics: Circle \& Ellipse}, where Gemini-2.5~\cite{google2025gemini25} scores 11.49 while Faire exceeds it by more than three times.
Crucially, these collapses align with categories where deductions are tightly state-dependent: the next reasoning step is valid only if the constructed configuration is exact.
The pale regions therefore diagnose a failure of distributional alignment: models reproduce the appearance of interleaved diagrams, yet the instantiated relations cannot reliably support subsequent reasoning.

In contrast, Faire forms a uniformly dark column across all categories, without localized degradation.
It maintains high scores on \emph{Functions \& Graphs} (70.93), \emph{Exponential/Log/Power Functions} (59.52), and \emph{Lines \& Coordinates} (59.11), demonstrating category-robust reasoning grounded in executable geometric states.
This consistency reflects functional alignment: construction is not a visual byproduct, but a reliable intermediate state actively used by the solver.
These results establish a methodological conclusion.
When interleaving is optimized only at the distributional level, increasing model scale does not prevent geometric collapse.
By contrast, enforcing functional alignment through post-generation verification enables a smaller model to systematically outperform much larger generalist MLLMs.
This shift, from imitating interleaving to internalizing construction as reasoning, defines the core advantage of Faire.

\section{Results on GGBench.}
\label{GGBench}
Table~\ref{tab:ggbench_results} reports stage-wise evaluation on GGBench~\cite{wei2025ggbench}, decomposing interleaved reasoning into \emph{Planning} (VLM-T), \emph{Middle Process} (VLM-I-Mid), and \emph{Final Result} (VLM-I-Res), together with pixel-level similarity metrics and an overall interleaved score (VLM-I).
This breakdown directly probes our central question: whether a model truly reasons through constructed geometric states, or merely arrives at plausible final outcomes.

Faire attains the strongest overall interleaved performance (VLM-I = 63.13), despite operating at a dramatically smaller scale than proprietary systems such as GPT-5 and Claude Sonnet~4.5.
The advantage is not marginal but structural, driven by a decisive lead in the \emph{middle process} score: Faire reaches 89.11, while the closest competitor remains below 78.
This near one-generation gap indicates that Faire is uniquely capable of sustaining step-by-step geometric state grounding, rather than skipping intermediate reasoning stages.

The contrast is particularly revealing for large generalist models.
GPT-5 and Claude Sonnet~4.5 exhibit strong planning and reasonable final outcomes, yet their intermediate scores fall well short of Faire, exposing a consistent process bottleneck.
These models often succeed by jumping from intent to answer, whereas Faire maintains verifiable state support at each reasoning step.
That an 8B model surpasses such large-scale systems underscores our main claim: functional alignment matters more than scale for interleaved geometric reasoning.

Pixel-level metrics further reinforce this conclusion.
Faire records a relatively low PSNR but achieves the highest SSIM across all models.
This is not a weakness but a signature of design choice.
PSNR rewards pixel-level smoothness and cosmetic rendering, while SSIM captures structural fidelity.
Faire deliberately prioritizes geometric topology and relational correctness over visual polish, trading pixel aesthetics for structural truth.
End-to-end image models, by contrast, can achieve visually pleasing outputs without reliable intermediate grounding, which limits their interleaved reasoning scores.

Overall, the GGBench results provide converging evidence for our thesis.
Optimizing interleaved reasoning requires enforcing how constructions function within the reasoning process, not merely scaling models or matching output appearance.
By turning intermediate constructions into reliable working states, Faire achieves a qualitative shift in geometric reasoning behavior that larger generalist models fail to realize.

\section{More \textbf{Aha Moment} Cases: From Format Imitation to Functional Alignment}
\label{sec:qualitative}
\begin{figure}[ht!]
  \centering
  \includegraphics[width=\linewidth]{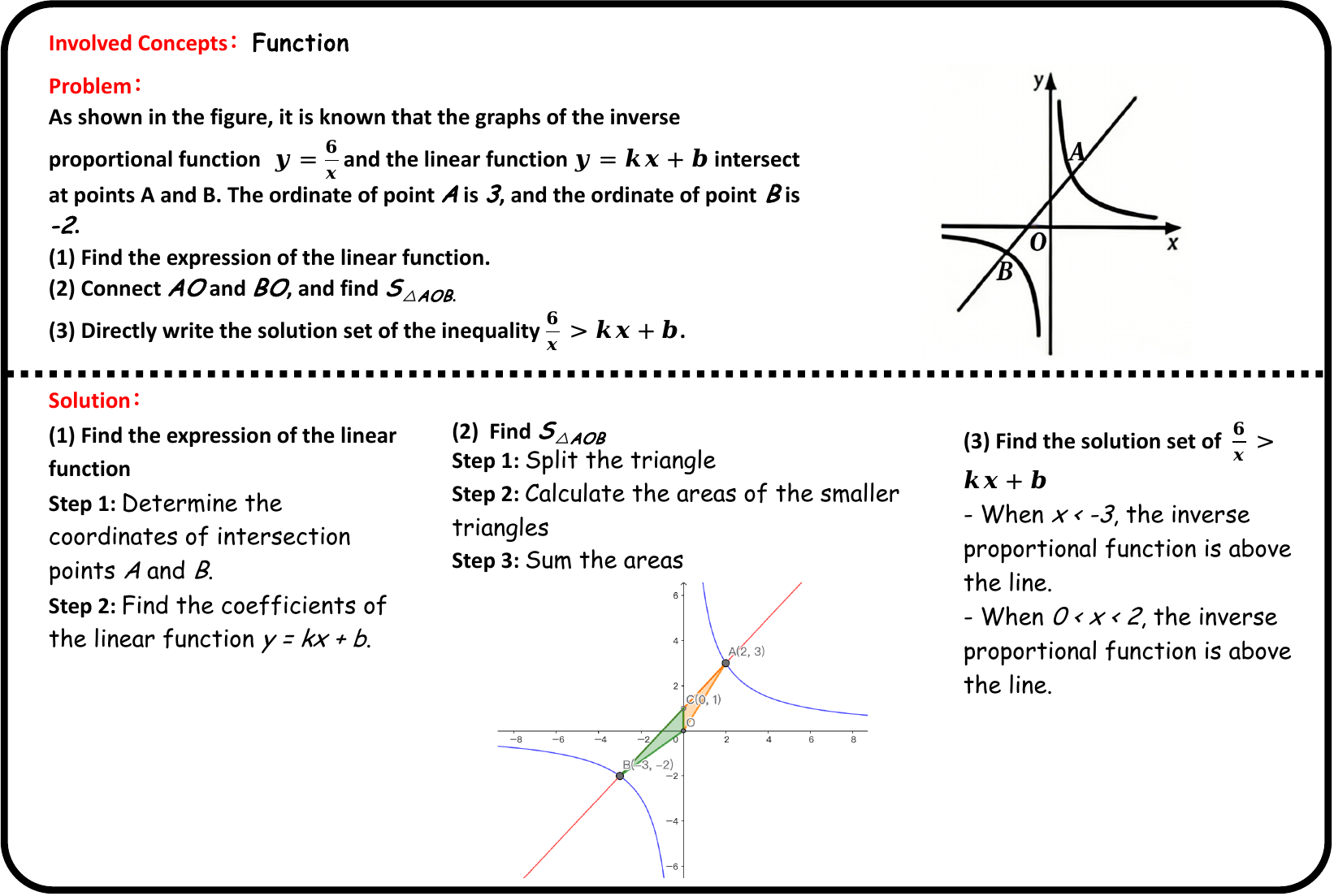}
  \caption{\textbf{Interleaved function visualization (line + reciprocal curve).}
  Faire constructs both graphs in the same coordinate system and marks the key points needed by the reasoning, so the option decision is supported by a checkable geometric state.}
  \label{fig:case_func_hyperbola}
  \vspace{-2mm}
\end{figure}

\begin{figure}[t]
  \centering
  \includegraphics[width=\linewidth]{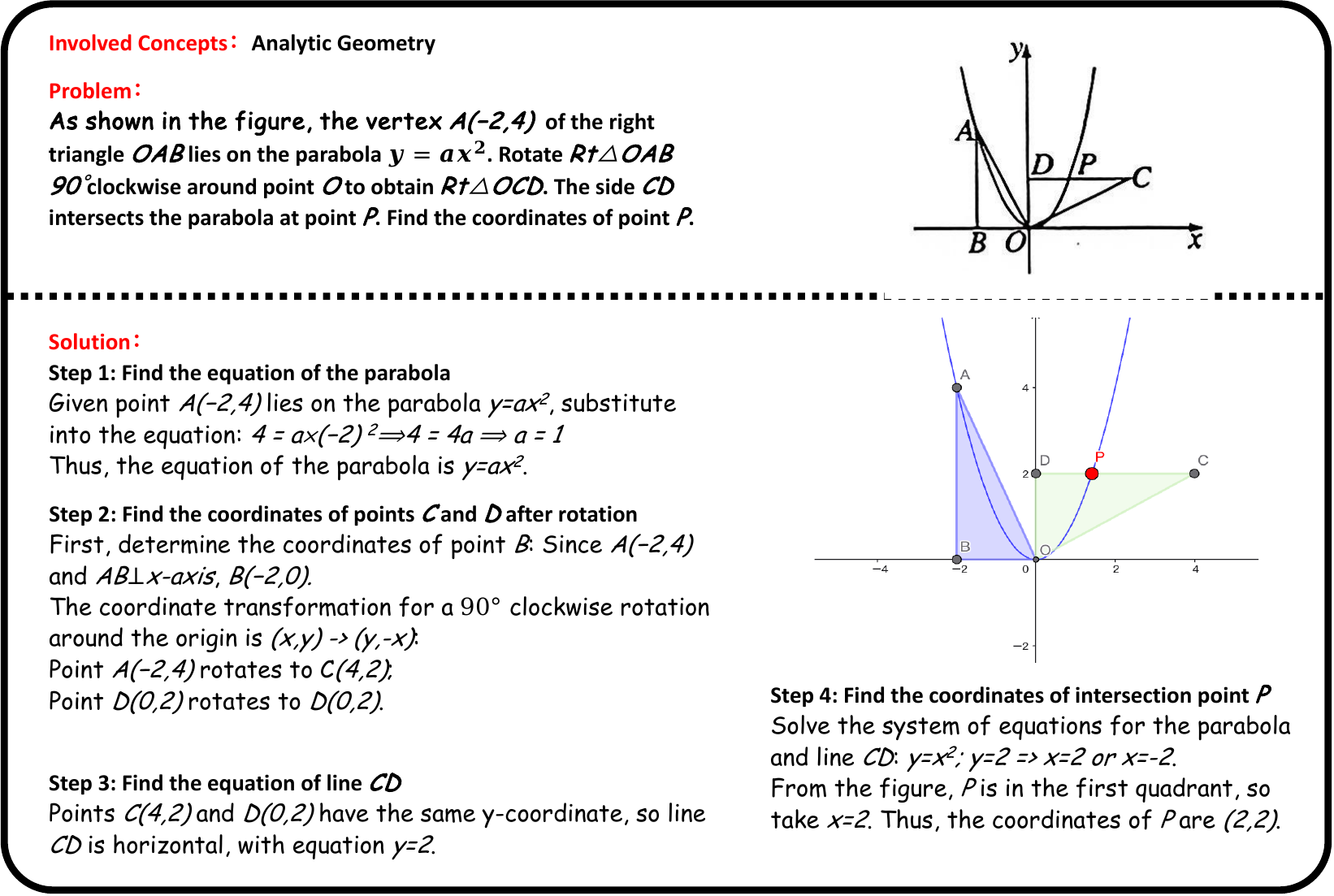}
  \caption{\textbf{Analytic-geometry construction with projections.}
  The diagram materializes auxiliary points and projection relations, turning abstract constraints into explicit incidences that the subsequent deduction can rely on.}
  \label{fig:case_parabola_projection}
  \vspace{-2mm}
\end{figure}

\begin{figure}[t]
  \centering
  \includegraphics[width=\linewidth]{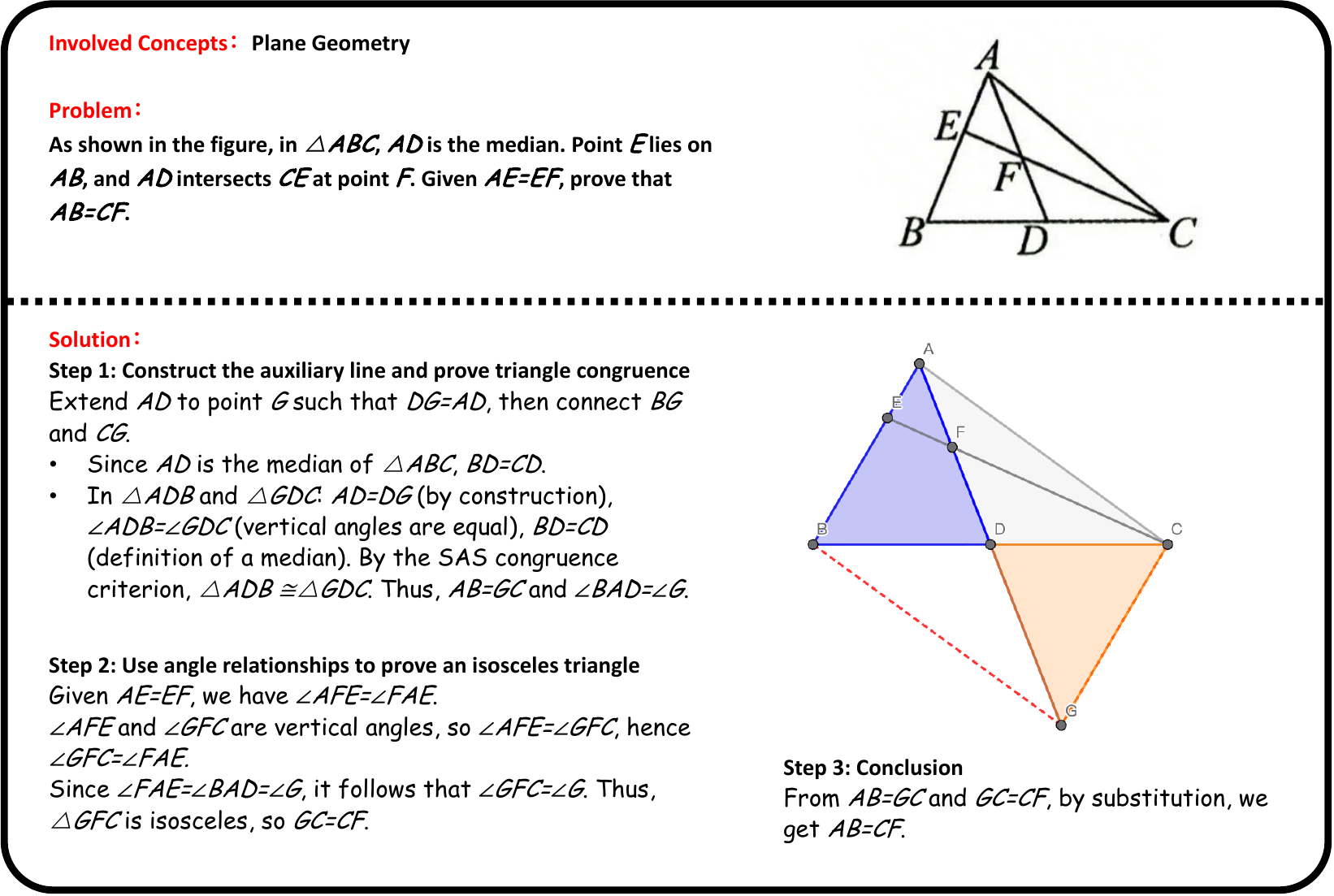}
  \caption{\textbf{Plane-geometry construction via auxiliary intersections.}
  The construction exposes hidden structure (intersections and induced sub-triangles), providing a concrete state for validating proportional or area arguments.}
  \label{fig:case_triangle_aux}
  \vspace{-2mm}
\end{figure}

\begin{figure*}[t]
  \centering
  \includegraphics[width=\textwidth]{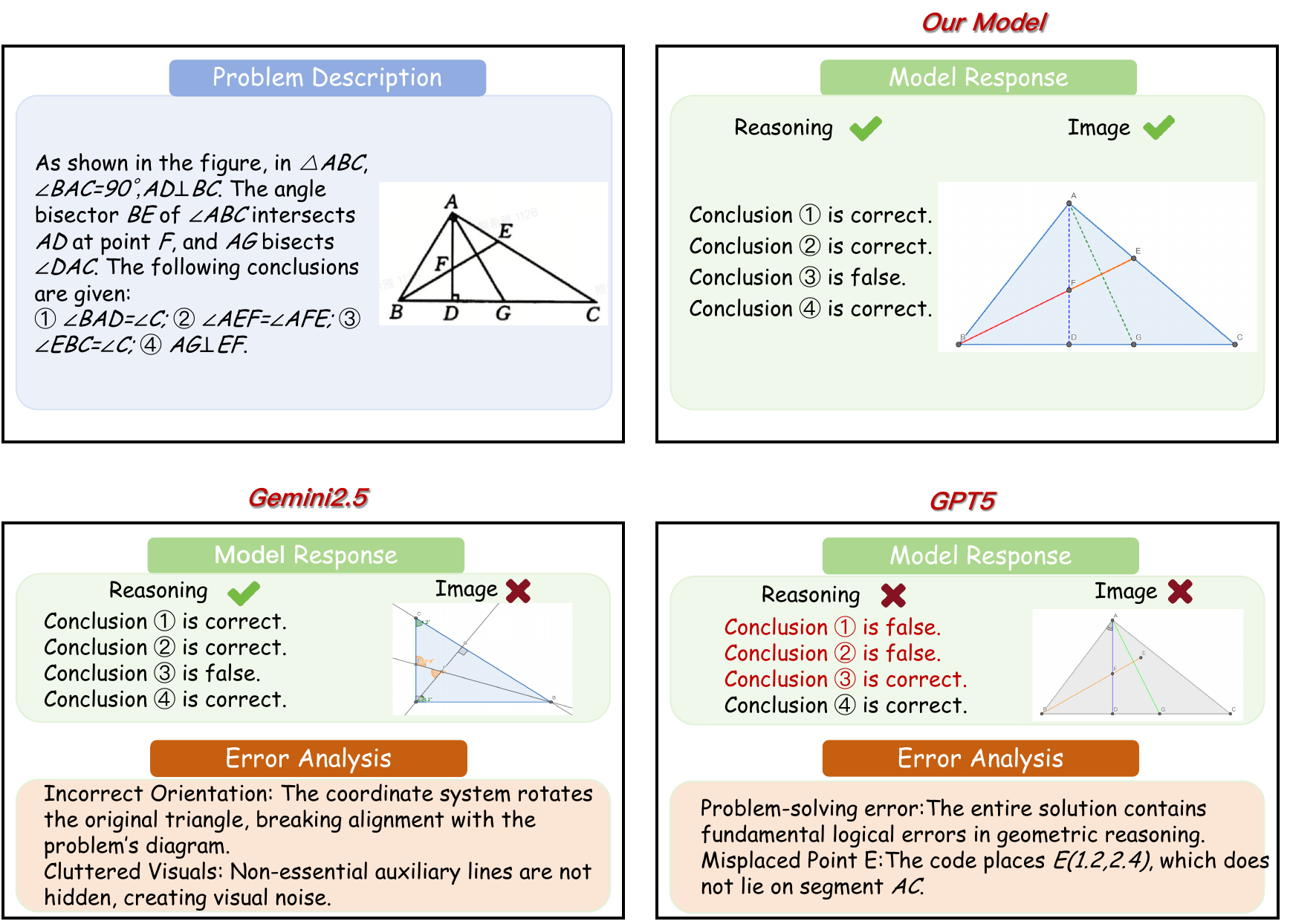}
  \caption{\textbf{Reasoning–diagram consistency comparison on a judgement problem.}
  Faire keeps deductions and construction aligned to the same geometric state; baselines may produce answer-correct text with diagram drift (orientation/clutter) or fail in both reasoning and construction.}
  \label{fig:case_mc_compare}
  \vspace{-2mm}
\end{figure*}

% \begin{figure*}[t]
%   \centering
%   \includegraphics[width=\textwidth]{images/demo5.pdf}
%   \caption{\textbf{SFT paradox in interleaved proof.}
%   Interleaved SFT can imitate the alternation format yet break causal coupling, leading to a wrong proof step and a mismatched construction that cannot support verification.}
%   \label{fig:case_proof_sft}
%   \vspace{-2mm}
% \end{figure*}

% \begin{figure*}[t]
%   \centering
%   \includegraphics[width=\textwidth]{images/demo6.pdf}
%   \caption{\textbf{Another interleaved SFT failure.}
%   A local logical mistake (equality misinterpreted as midpoint) propagates into the diagram, illustrating how weak text–diagram coupling makes interleaving a burden rather than a tool.}
%   \label{fig:case_proof_sft_dup}
%   \vspace{-2mm}
% \end{figure*}

\begin{figure*}[t]
  \centering
  \includegraphics[width=\textwidth]{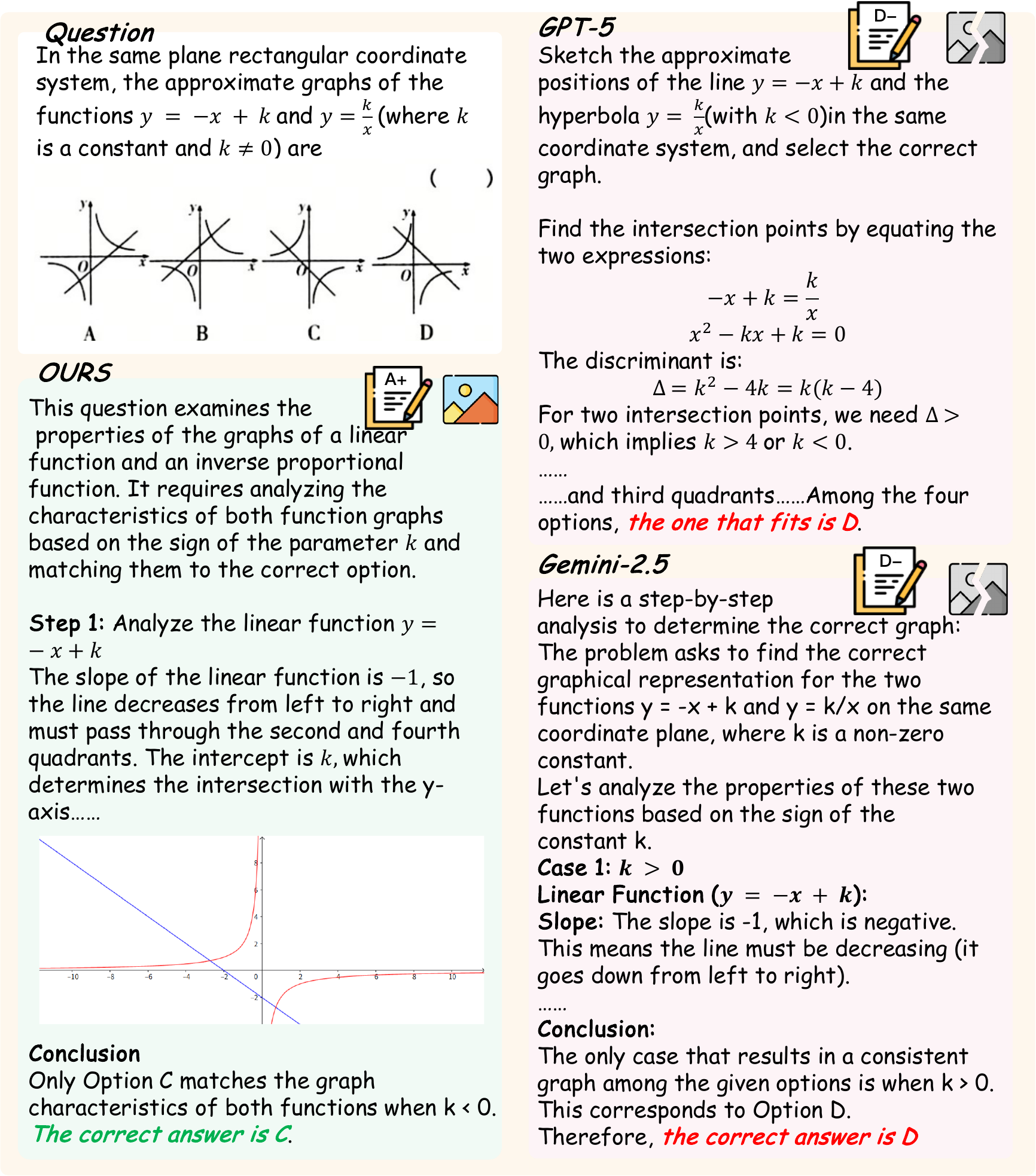}
  \caption{\textbf{Function graph selection: grounded structure versus ungrounded calculation.}
  Faire anchors the decision in a state-faithful plot under the given sign constraints, while large generalist models may compute extensively yet still select an option inconsistent with the intended configuration.}
  \label{fig:case_function_choice}
  \vspace{-2mm}
\end{figure*}
% We analyze five representative cases (Figures~\ref{fig:case_func_hyperbola}--\ref{fig:case_function_choice}) to reveal how interleaved reasoning succeeds or fails in practice.
% Across all examples, a clear pattern emerges: models trained with interleaved SFT can reproduce the \emph{appearance} of diagrammatic reasoning, but fail to maintain a causal dependency between the constructed diagram and the deductions that follow.
We analyze five representative cases (Figures~\ref{fig:case_func_hyperbola}--\ref{fig:case_function_choice}) to illustrate how interleaved reasoning succeeds or fails in practice.
Across these examples, Faire consistently treats the diagram as an executable intermediate state, using construction feedback to ground and constrain subsequent deductions.
In contrast, when compared with strong general-purpose models (e.g., Gemini and GPT), failures typically arise from weaker text–diagram coupling: diagrams are produced as illustrative artifacts rather than as binding states that actively guide the reasoning process.

In contrast, Faire consistently treats the diagram as an executable intermediate state, using it to support and constrain subsequent reasoning.

In problems where correct solutions hinge on precise geometric instantiation, such as function intersections, projections, and auxiliary constructions (Figures~\ref{fig:case_func_hyperbola}, \ref{fig:case_parabola_projection}, and \ref{fig:case_triangle_aux}), Faire produces diagrams whose geometric relations directly justify the next step in the solution.
Intersections, alignments, and auxiliary points are not merely visual cues but function as verifiable evidence.
By comparison, baseline models often generate visually plausible diagrams that fail to encode the relations their own reasoning implicitly assumes, causing downstream deductions to rest on unsupported geometric states.

A more revealing failure mode appears when symbolic answers are correct but constructions are not.
Figure~\ref{fig:case_mc_compare} shows that Gemini-2.5 reaches the correct logical conclusions, yet its generated diagram violates orientation and connectivity constraints, rendering it unusable as a proof artifact.
GPT-5 fails more fundamentally, producing both incorrect reasoning and invalid constructions.
These cases demonstrate that correctness of the final answer does not imply correctness of the constructed geometric state.
Only Faire satisfies symbolic reasoning, geometric construction, and visual consistency simultaneously.

% The most systematic breakdown is observed under interleaved SFT.
% Figures~\ref{fig:case_sft_failure_1} expose a recurring pattern: the model alternates between text and diagrams as instructed, yet misinterprets the semantic role of constructions.
Equal-length relations are mistaken for midpoint relations, auxiliary lines are introduced without the constraints they are meant to encode, and errors propagate across steps.
Here, interleaving becomes a liability rather than an aid, confirming that SFT optimizes surface alternation instead of functional dependency.

Faire resolves these failures by enforcing post-generation verification at each step.
Every constructed diagram is checked against the intended geometric relations before it is used for further reasoning.
As a result, incorrect constructions are rejected early, preventing silent error accumulation.
The diagram is no longer a decorative output but a proof-carrying state that constrains the reasoning process.

Finally, Figure~\ref{fig:case_function_choice} shows that functional grounding remains critical even when algebraic manipulation is correct.
Both GPT-5 and Gemini-2.5 derive valid intersection conditions, yet select the wrong graph due to a failure to reason globally about quadrant structure.
By explicitly constructing and inspecting the geometric configuration, Faire arrives at the correct choice.
This highlights a broader point: interleaved reasoning succeeds not by adding diagrams, but by making constructions functionally indispensable to deduction.

\end{document}

%% file: macro.tex
\usepackage{natbib}
\usepackage{latexsym}

\usepackage{url}
\usepackage{amssymb}
\usepackage[utf8]{inputenc}
\usepackage{microtype}
\usepackage{booktabs}
\usepackage{pifont} 
\usepackage{multirow}
\usepackage{makecell}
\usepackage{paralist}
\usepackage{xspace}
\usepackage{color}
\usepackage{xcolor}
\usepackage{colortbl}
\usepackage{adjustbox}
\usepackage{hyperref} 
\usepackage[edges]{forest}
\usepackage{tikz} 
\usepackage{caption}
\usepackage{amsfonts}

\hypersetup{
    colorlinks,
    linkcolor={blue!80!black},
    citecolor={blue!80!black},
}
\tikzset{
    root/.style =             {align=center, text width=1cm, rounded corners=3pt, line width=0.3mm, fill=gray!10, draw=gray!80, font=\small},
    demographic/.style =         {align=center, text width=1.8cm, rounded corners=3pt, line width=0.3mm, fill=blue!10, draw=blue!80, font=\footnotesize},
    demographic_work/.style =    {align=center, text width=10cm, rounded corners=3pt, line width=0.3mm, fill=blue!10, draw=blue!0, font=\footnotesize},
    character/.style =         {align=center, text width=1.8cm, rounded corners=3pt, line width=0.3mm, fill=red!10, draw=red!80, font=\footnotesize},
    character_work/.style =    {align=center, text width=10cm, rounded corners=3pt, line width=0.3mm, fill=red!10, draw=red!0, font=\footnotesize},
    personalization/.style =           {align=center, text width=1.8cm, rounded corners=3pt, line width=0.3mm, fill=cyan!10, draw=cyan!80, font=\footnotesize},
    personalization_work/.style =      {align=center, text width=10cm, rounded corners=3pt, line width=0.3mm, fill=cyan!10, draw=cyan!0, font=\footnotesize},
    risk/.style =         {align=center, text width=1.8cm, rounded corners=3pt, line width=0.3mm, fill=orange!10, draw=orange!80, font=\footnotesize},
    risk_work/.style =    {align=center, text width=10cm, rounded corners=3pt, line width=0.3mm, fill=orange!10, draw=orange!0, font=\footnotesize},
}

\newcommand{\cmark}{\ding{51}}
\newcommand{\xmark}{\ding{55}}

\usepackage{CJK}

%% file: sections/000abstract.tex
Solving complex geometric problems inherently requires \textit{interleaved reasoning}: a tight alternation between constructing diagrams and performing logical deductions. Although recent Multimodal Large Language Models (MLLMs) have demonstrated strong capabilities in visual generation and plotting, we identify a counter-intuitive and underexplored phenomenon. Naively applying Supervised Fine-Tuning (SFT) on interleaved plot–solution data leads to a substantial degradation in reasoning performance compared to text-only baselines. We argue that this failure stems from a fundamental limitation of SFT, which primarily induces \textit{distributional alignment}: the model learns to reproduce the surface format of interleaved plotting but fails to internalize the causal dependency between the generated plot and reasoning steps. To overcome this limitation, we propose Faire (\textbf{F}unctional \textbf{a}lignment for \textbf{i}nterleaved \textbf{re}asoning), a reinforcement learning framework that enforces three casual constraints to move beyond superficial imitation toward \textit{functional alignment}. Extensive experiments show that Faire induces a qualitative shift in model behavior in which the plotting is effectively internalized, yielding competitive performance on challenging geometric reasoning benchmarks.

% We introduce \method, capable of reasoning through thinking before responding, resulting in improved performance on a wide range of benchmarks.
% \method achieves $86.7$ on AIME $2024$, $55.0$ on Codeforces and $77.3$ on GPQA, demonstrating excellent reasoning abilities in STEM and coding.
% Beyond reasoning tasks, the method demonstrates notable generalization across diverse domains. For instance, it surpasses DeepSeek R1 by 8\% in win rate on non-reasoning tasks, indicating its broader applicability.
% Compared to other state-of-the-art reasoning models, \method is a Mixture-of-Experts (MoE) model with a relatively small size, featuring 20B activated and 200B total parameters.
% As part of our effort to assess generalized reasoning, we develop two internal benchmarks, BeyondAIME and Codeforces, both of which will be publicly released to support future research. Model trial link: \url{https://www.volcengine.com/experience/ark}.

\date{January 29, 2026}
% \Correspondence{January 29, 2026}
\correspondence{\email{tancheng@pjlab.org.cn}, \email{weijingxuan20@mails.ucas.edu.cn}}

%% file: sections/010intro.tex
\section{Introduction}

Solving complex geometric problems is rarely a linear trajectory of purely textual deduction~\cite{mouselinos2024beyond,xu2024geo}. Instead, it requires interleaved reasoning, a dynamic interplay where experts iteratively construct visual diagrams to ground their logical derivations~\cite{akter2024self,ning2025gns}. Within this cognitive loop, the visual plot serves not merely as a static illustration, but as an essential cognitive scaffold: it exteriorizes working memory, makes latent spatial constraints explicit~\cite{qicogcom,kumar2025diagramir}, and actively steers the subsequent deductive path. With the rapid evolution of Multimodal Large Language Models (MLLMs), integrating such \textit{thinking-in-plots} capabilities has emerged as a frontier pursuit~\cite{mathvista,zhuang2025math}. The ideal MLLM must therefore transcend the role of a passive calculator, evolving into a comprehensive solver~\cite{lilarge,wang2024measuring} that seamlessly interleaving code-based plotting with textual reasoning to achieve superior problem-solving accuracy.

\begin{figure}[ht]
    \centering
    \includegraphics[width=0.7\textwidth]{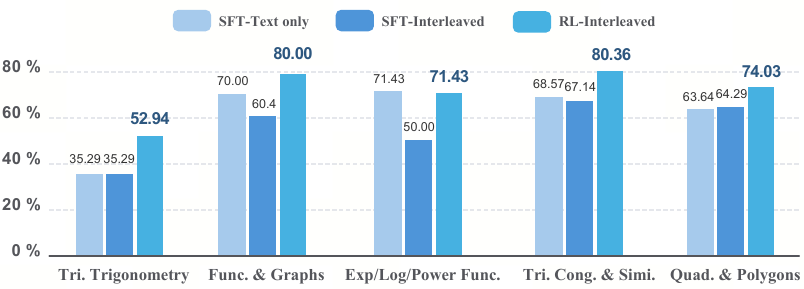}
    \caption{Illustration of challenges of \textit{geometric interleaved reasoning} on several sub-tasks between SFT-Text only and SFT-Interleaved, which can only be tackled by RL post-training.}
    \label{fig:intro}
    \vspace{-1.0em}
\end{figure}

However, existing training paradigms reveal a counterintuitive limitation~\cite{zhangmultimodal,wang2024t,shao2024visual}. While Supervised Fine-Tuning (SFT) on high-quality interleaved plot–solution data is expected to improve reasoning by providing richer multimodal context~\cite{gao2025interleaved,gu2025thinkmorph,zheng2023ddcot}, our empirical results show the opposite effect. As shown in Figure~\ref{fig:intro}, naively applying SFT to such interleaved traces leads to a significant performance drop compared to text-only baselines. Rather than acting as a scaffold, the insertion of plotting steps disrupts the reasoning process by forcing the model to alternate between logical deduction and code generation.
% This context switching breaks the continuity of the reasoning, making it difficult for the model to resume coherent reasoning after plotting.
We argue that this failure stems from a fundamental limitation of SFT. SFT primarily induces distributional alignment by minimizing divergence from the training data, causing the model to imitate the surface form of interleaved reasoning, such as when to generate visual outputs, without internalizing the causal dependency that makes visual construction necessary for correct inference. As a result, visual generation is treated as a noisy pathway rather than an integral component of problem solving, which ultimately leads to the observed degradation in performance.

To bridge this gap in geometric reasoning, we posit that Reinforcement Learning (RL) is necessary to move beyond superficial \textit{distributional alignment} toward \textit{functional alignment}. We propose Faire (\textbf{F}unctional \textbf{a}lignment for \textbf{i}nterleaved \textbf{re}asoning), a RL–based framework that explicitly enforces functional alignment through preference optimization. Central to Faire is a tri-perspective verifier system that imposes strict causal constraints on the generation process. Specifically, an objective verifier validates geometric correctness through executable programs; a subjective verifier ensures visual perceptibility and interpretability via feedback from a vision–language model; and a semantic verifier enforces consistency between the generated reasoning, the constructed diagram, and the problem statement. 

The functionally aligned reward mechanism mines the long-tail of the generation distribution, reinforcing those rare outlier trajectories where the diagram actively facilitates deduction. Faire not only reverses the negative transfer observed in SFT but also validates our theoretical hypothesis regarding functional alignment, ultimately establishing state-of-the-art performance on challenging geometric benchmarks where the model genuinely internalizes the interleaving thinking-in-plots paradigm.

%% file: sections/020related.tex
\section{Related Work}

\subsection{Multimodal Geometry Reasoning}

Prior work on multimodal geometry problem solving can be broadly categorized into data-centric and perception-driven approaches.
Representative efforts range from early neural solvers that align diagrams with textual representations~\cite{ijcai2023p376}, to instruction-tuned MLLMs and large-scale geometry datasets designed to enhance multimodal alignment and reasoning diversity~\cite{ICLR2025_09afabe3,wu-etal-2023-conic10k,10.1145/3696409.3700262,10960701}.
Related studies further emphasize symbol-aware modeling to better ground geometric entities across text and vision~\cite{10.1145/3581783.3612570}.
While these methods enhance perceptual grounding and data efficiency, their reasoning are weakly constrained, with implicit intermediate states.

In contrast, logic-centric approaches emphasize rigor and verifiability through neuro-symbolic or execution-based formulations, including language-guided symbolic systems~\cite{trinh2024solving,anonymous2025euclidomni}, formally structured representations and benchmarks~\cite{zhang2024formal,wei2025geoint}, executable code generation for geometry solving~\cite{sharma-etal-2025-geocoder}, and formally verified supervision engines~\cite{fu2025trustgeogen,wu2025nesygeo}.
Interleaved paradigms that couple reasoning with explicit drawing actions have been proposed~\cite{Wu2025ReinforcingSR}.
However, relying solely on supervised learning for such integration often leads to negative transfer, where models mimic the interaction without capturing its causal utility.
% However, tightly coupling abstract reasoning with low-level geometric execution can introduce cognitive interference, where maintaining execution constraints competes with high-level planning.
% This tension motivates approaches that decouple reasoning from plotting, enabling precise visualization and rigorous logic to reinforce each other without entanglement.

\subsection{Interleaved Reasoning and Unified Generation}

Recent multimodal research moves beyond static VQA toward \emph{interleaved} reasoning traces and unified image-text generation. Representative efforts include interleaved chains of thought and progressive interleaved benchmarks~\cite{gao2025interleaved,du2025easy}, visual evidence chaining with region-level grounding~\cite{shao2024visual}, and reflection-style interleaving for text-to-image generation~\cite{huang2025interleaving}.
In parallel, tool- and program-augmented paradigms translate perception into executable or inspectable intermediate states via tool orchestration, modular programs, and iterative search~\cite{yang2023mm,suris2023vipergpt,xu2025mars2,wang-etal-2025-visuothink,ding2025arm}.
On the modeling side, unified architectures seek to handle interleaved understanding and generation within a single system, spanning unified I/O alignment~\cite{koh2023generating}, unified tokenization~\cite{ge2024making}, single-transformer unification and autoregressive interleaved generation~\cite{xieshow,kouorthus}, as well as multimodal-prompted and interactive generation~\cite{huang2025wegen}.

To support this ecosystem, recent benchmarks evaluate open-ended interleaved outputs from multiple dimensions~\cite{zhou2025opening,liu2024holistic}, while challenge-style testbeds provide a dynamic substrate for tracking rapidly evolving multimodal reasoning systems~\cite{xu2025mars2}, and RL-based post-training further improves interleaved generation without requiring massive supervised trajectories~\cite{nietowards}.
However, existing paradigms prioritize the form of interleaved generation over its functional utility, often resulting in the negative transfer we observe in geometric tasks.
% However, integrating reasoning, acting, and generation into a monolithic interleaved stream often induces capability conflict and attribution noise; recent evidence demonstrates that decoupling visual encoding paths mitigates interference between understanding and generation objectives~\cite{wu2025janus}.
% In the context of geometry, this tension becomes particularly acute: tightly interleaving high-level mathematical reasoning with low-level visual construction or plotting can obscure intermediate logical states, motivating traceability-oriented designs that explicitly separate reasoning from geometric construction while preserving their alignment.

%% file: sections/030data.tex
\section{Method}

\label{sec:method}

\begin{figure*}[t]
    \centering
    \includegraphics[width=1.0\textwidth]{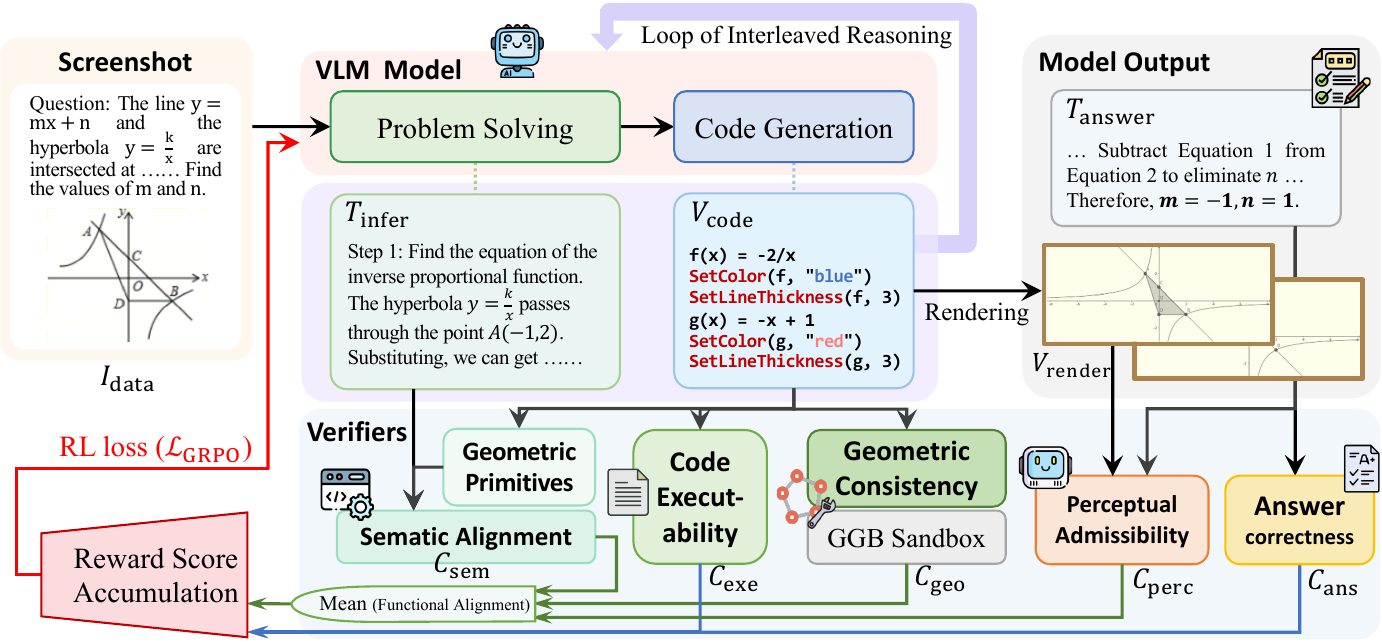}
    \caption{Illustration of Faire framework.
    The model generates a reasoning trace and GeoGebra code from a geometry problem.
    As for reward designs, a gated reward enforces answer correctness $C_{ans}$ and code executability $C_{exe}$, then aggregates perceptual $C_{perc}$, semantic $C_{sem}$, and formal verification signals $C_{geo}$.}
    \label{fig:framework}
    \vspace{-1.0em}
\end{figure*}

\subsection{Preliminaries}
\label{sec:method_setup}

We formally define the task of \textit{geometric interleaved reasoning}. Unlike standard text-only mathematical reasoning, where the solution is a continuous stream of logical tokens, interleaved reasoning requires the model to dynamically alternate between linguistic deduction and executable visual construction. Given a geometric problem $X$ that may include a problem statement and an initial diagram, the goal is to generate a solution trajectory $\tau$:
\begin{itemize}
    \vspace{-1mm}
    \item \textbf{Text-only reasoning}: The trajectory is a homogeneous sequence of text tokens $\tau_{cot} = [t_1, t_2,...,t_N]$, where each step is generated conditionally on the prior linguistic context, following $P(t_k|t_{<k},X)$.
    \vspace{-1mm}
    \item \textbf{Interleaved reasoning}: The trajectory is a heterogeneous sequence of reasoning steps and plotting actions:
    \begin{equation}
        \tau_{int} = [(t_1, c_1), (t_2, c_2),...,(t_k, c_K)].
    \end{equation}
\end{itemize}
Here, $t_k$ represents a textual reasoning block, and $c_k$ represents an executable GeoGebra code block that generates a visual state $v_k=\mathrm{Render}(c_k)$. The defining distinction between the two paradigms lies in the \textit{conditional dependency}. In interleaved reasoning, the next textual step $t_{k+1}$ is conditioned not only on the preceding text, but also on the accumulated visual context:
\begin{equation}
    P(t_{k+1}|t_{1:k},v_{1:k},X).
\end{equation}
When properly aligned, the visual state $v_k$ serves as an external memory that clarifies latent constraints and facilitates subsequent deduction. Otherwise, the text-to-code transition introduces interference, and the generated diagram degrades into a nuisance signal that fragments the reasoning process.

\subsection{Distributional Alignment Leads to Failure}

To formally analyze the divergence between SFT and RL, we abstract the sequential trajectory into a probabilistic graphical model. Let $T=\{t_1,...,t_k\}$ denote the aggregate textual rationale, and $V=\{v_1,...,v_k\}$ denote the aggregate visual artifacts generated throughout the reasoning process. The reasoning process is a joint distribution over the textual rationale $T$, the visual artifact $V$, and the final answer $Y$, conditioned on the problem context $X$.

SFT optimizes the policy $\pi_\theta$ to minimize the Kullback-Leibler (KL) divergence from the data distribution $\mathcal{D}$:
\begin{equation}
    \mathcal{L}_{SFT}(\theta) = \mathbb{E}_{\tau \sim \mathcal{D}} [-\log \pi_\theta(T, V, Y|X)].
\end{equation}
The data distribution implies a causal structure where the expert generates $V$ specifically to simplify $Y: V \rightarrow Y$. However, SFT merely learns the conditional likelihoods:
\begin{equation}
\begin{aligned}
    \log \pi_\theta(\tau|X) &= \underbrace{\log \pi_\theta(T_1|X)}_{\text{Premise}} + \underbrace{\log \pi_\theta(V|T_1, X)}_{\text{Plotting}} \\ &+ \underbrace{\log \pi_\theta(T_2, Y|T_1, V, X)}_{\text{Deduction}}
\end{aligned}
\end{equation}
We identify two fundamental issues under the SFT regime:

\textbf{(1) Superficial correlation} SFT minimizes the prediction error of $V$, i.e., minimizing conditional entropy $H_{\pi}(V | T_1, X)$. The model learns to generate $V$ that looks like the training data (distributional alignment) but effectively treats $V$ and $Y$ as conditionally independent given strong textual priors:
\begin{equation}
    I_{\text{data}}(V; Y | T_1, X) - I_{\pi_\theta}(V; Y | T_1, X) \gg 0.
\end{equation}
Here, $I(\cdot;\cdot|\cdot)$ is conditional mutual information. The trained policy underestimates the dependency between $V$ and $Y$.

\textbf{(2) Context fragmentation} When $\pi_\theta$ generates a hallucinated scaffold, it acts as a noisy channel. Instead of reducing the uncertainty of the answer, the inclusion of $V$ introduces a distraction penalty. We formalize this as an inequality in conditional entropy:
\begin{equation}
    H_{\pi_\theta}(Y | T_1, V, X) \ge H_{\pi_\theta}(Y | T_1, X) + \delta,
\end{equation}
where $\delta > 0$ represents the cognitive load of reconciling the conflict between the text context $T_1$ and the noisy diagram $V$. Thus, SFT interleaved data increases the entropy of the solution space rather than collapsing it.
% Executing the accumulated code blocks yields a concrete diagram state, which externalizes the geometry that the subsequent reasoning step is expected to rely on.
% This instantiates a dual process: reasoning proposes symbolic intentions, while construction performs concrete actions that materialize those intentions.
% In the ideal regime, diagram construction is not a standalone output; it functions as the working memory of the solution.

% This idealized interaction, however, is not realized by supervised fine-tuning.
% Although interleaved SFT faithfully reproduces the surface pattern of alternating text and code, it fails to enforce a causal dependency between what is constructed and what is reasoned next.
% The model learns \emph{how} to interleave, but not \emph{why}.
% As a result, plotting becomes a burden rather than a tool: it consumes capacity without reliably supporting downstream deductions, and can even degrade reasoning performance relative to text-only baselines.

% To restore this missing dependency, we optimize the policy with reinforcement learning.
% By assigning credit only when constructed states are executable and verifiably consistent with the intended reasoning, RL enforces \emph{functional alignment} between deduction and construction.
% This training regime leads to the \textbf{Aha Moment}, where diagram construction ceases to be a decorative byproduct and instead becomes functionally coupled to subsequent reasoning steps.
\subsection{Functional Alignment for Causal Dependency}

To reverse the degradation caused by distributional alignment, we propose \textit{functional alignment}. Unlike SFT which treats the visual artifact $V$ as a supervised target, we model $V$ as a \textit{latent causal mediator} in the reasoning graph $T \to V \to Y$. RL optimizes a policy that instantiates $V$ to serve as an information bridge between the textual premise $T$ and the answer $Y$. For the causal path $T \xrightarrow{\text{construct}} V \xrightarrow{\text{observe}} Y$ to be valid, the mediator $V$ must satisfy a set of necessary structural constraints. We define a Tri-perspective verification system $\mathcal{V}$ that formalizes these constraints, ensuring the structural integrity of the reasoning graph.

\paragraph{Necessary conditions for causal mediation}
We posit that $V$ is a functional mediator if and only if it satisfies three independent conditions: \textit{Geometric Consistency} ($\mathcal{C}_{geo}$), \textit{Perceptual Admissibility} ($\mathcal{C}_{perc}$), and \textit{Semantic Alignment} ($\mathcal{C}_{sem}$).

\noindent\textbf{(1) Node Integrity $\mathcal{C}_{geo}$}: We require that the mediator $V$ exists as a well-defined mathematical object. Concretely, the generated code $c$ must be executable and not violate any axioms implied by the problem $X$. We formalize it via the \textit{objective verifier}, which executes the code and checks against a set of hard constraints $\mathcal{K}_X$ derived from $X$:
\begin{equation}\mathcal{C}_{geo}(c, X) = \mathbb{I}(c \in \Omega_{valid}) \cdot \prod_{k \in \mathcal{K}_X} \mathbb{I}(\text{Satisfies}(c, k)),
\end{equation}
where $\Omega_{valid}$ denotes the space of syntactically valid programs, guaranteeing the ontological validity of the mediator.

\noindent\textbf{(2) Egress Validity $\mathcal{C}_{perc}$}: The information contained in $V$ must be transmittable to the solver. Even if $V$ is mathematically perfect, rendering artifacts (e.g., occlusion, extreme scale, label overlap) act as channel noise, blocking the edge $V \to Y$. We formalize this via the \textit{subjective verifier} using a MLLM as a judge function $J_{mllm}$:
\begin{equation}
\mathcal{C}_{perc}(v) = \mathbb{I}(J_{mllm}(v) > \tau_{perc}).
\end{equation}
Here, $\tau_{perc}$ is a threshold for visual clarity. Satisfying $\mathcal{C}_{perc}$ ensures that the channel capacity is sufficient.

\noindent\textbf{(3) Ingress Validity $\mathcal{C}_{sem}$}: The mediator $V$ must be causally downstream of the specific intent in $T$. A random valid figure (satisfying $\mathcal{C}_{geo}$ and $\mathcal{C}_{perc}$) that ignores the current reasoning plan implies a broken edge $T \nrightarrow V$ (i.e., mutual information $I(T; V) \approx 0$).We formalize this via the \textit{semantic verifier}. Let $\phi(c)$ be the set of geometric primitives extracted from code, and $\psi(T)$ be the set of geometric intents extracted from text. We require logical entailment:
\begin{equation}
\mathcal{C}_{sem}(c, T) = \mathbb{I}(\psi(T) \subseteq \phi(c)).
\end{equation}
This condition enforces that the diagram is not just a valid figure, but the specific figure necessitated by the textual logic, securing the ingress edge of the mediation.

\paragraph{Completeness of Verification} The proposed system is sufficient to establish functional alignment because it covers all components of the local causal graph: $\mathcal{C}_{sem}$ validates the input edge ($T \to V$), $\mathcal{C}_{geo}$ validates the node integrity ($V$), and $\mathcal{C}_{perc}$ validates the output edge ($V \to Y$):
\begin{equation}
    T \xrightarrow[\mathcal{C}_{sem}]{\text{construct}} \underbrace{V}_{\mathcal{C}_{geo}} \xrightarrow[\mathcal{C}_{perc}]{\text{observe}} Y.
\end{equation}
\vspace{-4mm}
\paragraph{Optimization} We integrate the proposed constraints into a unified dense reward function:
\begin{equation}
    R(V, Y) = \mathbb{I}(Y=Y^*) + \beta (\mathcal{C}_{geo} + \mathcal{C}_{perc} + \mathcal{C}_{sem}).
\end{equation}
Using GRPO~\cite{shao2024deepseekmath}, we update the policy to maximize the expectation of this causally-grounded reward:
\begin{equation}
\mathcal{L}_{GRPO} = \mathbb{E}_{\tau \sim \pi_\theta} \left[ \frac{\pi_\theta(\tau)}{\pi_{old}(\tau)} \hat{A}(\tau) \right] - \beta D_{KL}(\pi_\theta || \pi_{ref}).
\end{equation}
By explicitly rewarding these three conditions, Faire forces the model to internalize the causal structure of interleaved reasoning, ensuring that every generated plot functionally instrumental in deriving the final solution.

\begin{figure*}[t!]
    \centering
    \includegraphics[width=\linewidth]{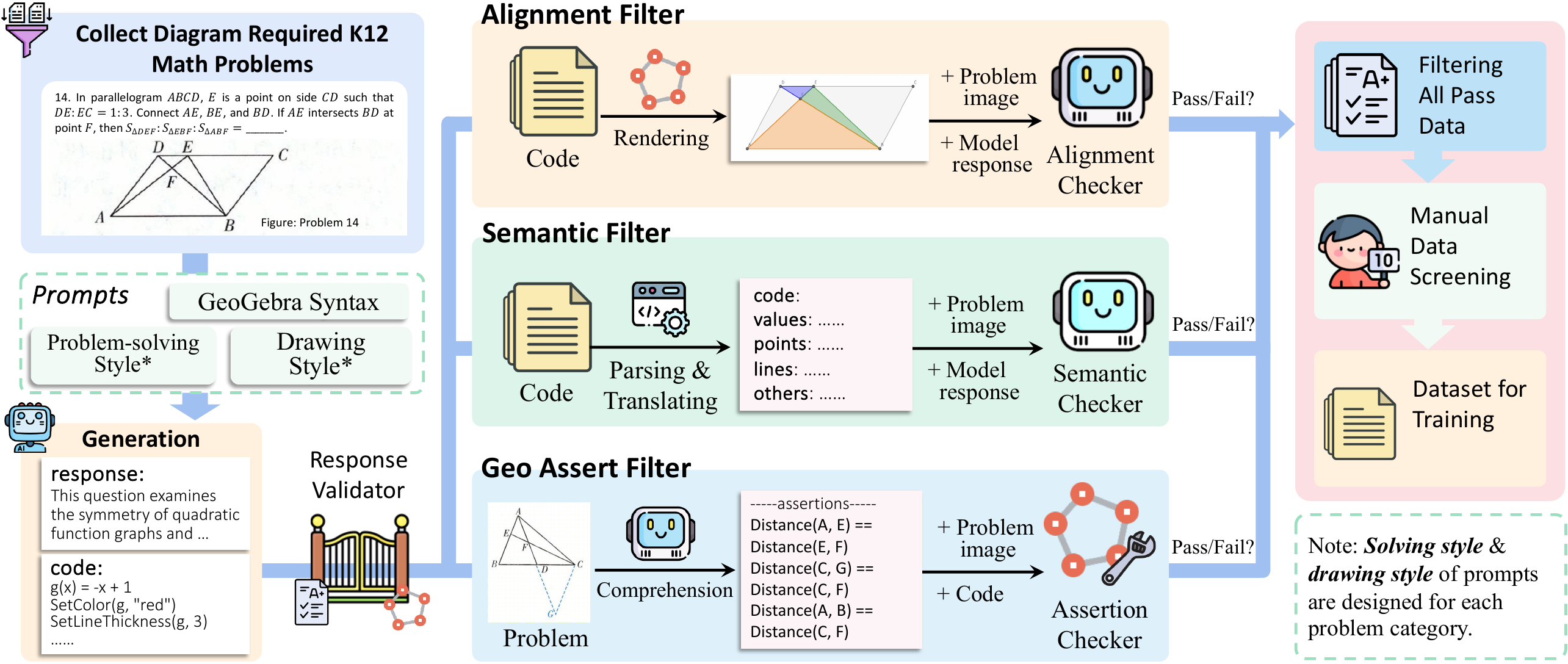}
    \caption{The data construction pipeline employing a tri-perspective verification mechanism—Visual Alignment, Semantic Consistency, and Geometric Assertion—to curate rigorous interleaved geometric reasoning samples.}
    \label{fig:pipeline}
    \vspace{-5mm}
\end{figure*}

\vspace{-2mm}

%% file: sections/040verify.tex
\section{Dataset}
\label{sec:verifier}

We introduce \textbf{Faire-Bench}, a benchmark for \emph{geometric interleaved reasoning} built with a synthesis-verification pipeline in Figure~\ref{fig:pipeline}.
Each instance couples stepwise deduction with an executable program whose execution reconstructs the geometric state that the next deduction is meant to rely on.
% We build Faire-Bench with a synthesis-verification pipeline, 
% that filters large-scale candidates through a three-stage funnel: perceptual usability, semantic intent alignment, and formal geometric validity.
% The resulting corpus provides deterministic, intent-oriented supervision beyond surface similarity, enabling a direct study of how RL unlocks the \textbf{Aha Moment}, where construction becomes a functional part of reasoning.
% Figure~\ref{fig:pipeline} provides an overview.
\vspace{-2mm}
\subsection{Dataset Construction}

\begin{table}[h]
\small
\centering
\vspace{3mm}
\caption{Consistent split-level statistics of the corpus.
% Percentages are reported within each split.
}
\vspace{-1mm}
\label{tab:dataset_stats}
\setlength{\tabcolsep}{4.2mm}
\begin{adjustbox}{max width=0.90\textwidth}
\begin{tabular}{lccc}
\toprule
 & SFT & RL & Eval \\
\midrule
Instances & 4643 & 2321 & 1025 \\
\midrule
Plane geometry & 51.5\% & 51.6\% & 54.7\% \\
Function & 29.4\% & 28.3\% & 27.4\% \\
Analytic geometry & 19.1\% & 20.1\% & 17.9\% \\
\midrule
Hard difficulty & 54.4\% & 54.8\% & 54.1\% \\
\midrule
$\geq$2 images & 39.6\% & 38.4\% & 43.6\% \\
\bottomrule
\end{tabular}
\end{adjustbox}
\vspace{-4mm}
\end{table}

\label{sec:dataset_construction}
We construct a K12 dataset of aligned solution traces and executable GeoGebra scripts via constrained synthesis and a verification funnel that yields \emph{All-Pass} samples.

\noindent\textbf{Taxonomy-driven selection}
We filter a large pool of K12 problems:
A lightweight taxonomy covers (i) function and analytic-geometry tasks that depend on axes, curves, and key points (e.g., intercepts, vertices, intersections, tangency points), and (ii) Euclidean construction tasks defined by primitives and relations (e.g., collinearity, perpendicularity, parallelism, tangency) with readable annotations.

\noindent\textbf{Constrained synthesis}
For each selected problem, Gemini~2.5 generates a stepwise solution trace and a GeoGebra script.
Prompts enforce three minimal constraints: (1) \textbf{Syntax}, the script is standalone and free of natural-language artifacts; (2) \textbf{Grounding}, reasoning steps explicitly reference the constructed objects; (3) \textbf{Drafting}, the diagram follows category-specific conventions (e.g., axes and key-point labels for plots, construction order for Euclidean tasks).
% We sample multiple candidates per problem to accommodate non-unique realizations.

\noindent\textbf{Hard filtering}
We discard incorrect candidates, then execute code in GeoGebra to ensure it constructs a valid geometric state and produces a non-degenerate render $I_{\text{render}}$.

\noindent\textbf{Verification funnel}
% Executable candidates still exhibit three distinct failure modes: a render can be perceptually unusable, a construction can deviate from the intended reasoning, or a visually plausible diagram can violate geometric constraints.
% We therefore apply three complementary filters that target these modes in order.
 We apply three complementary filters that target common failure modes:
(1) \textbf{Alignment Filter} compares $I_{\text{render}}$ against the problem diagram $I_{\text{prob}}$ (with \texttt{response} as context) to reject missing elements, disconnected auxiliary strokes, off-canvas renders, and severe label clutter.
(2) \textbf{Semantic Filter} parses \texttt{code} into $T_{\text{IR}}$ and verifies symbolic consistency between \texttt{response} and $T_{\text{IR}}$, preventing intent-shifted constructions that remain visually plausible.
(3) \textbf{Geo Assert Filter} synthesizes geometric assertions from the problem specification and evaluates them in the GeoGebra kernel, providing a deterministic truth verdict for key relations.
Only candidates that pass all filters enter expert review; the survivors form the final dataset.
% Only candidates that pass all filters enter expert review for notation and presentation edge cases; the survivors form the final dataset.

\subsection{Dataset analysis}
\label{sec:dataset_analysis}

\noindent\textbf{Split consistency and visual context}
Our corpus contains $7{,}989$ instances, partitioned into an SFT ($4{,}643$)/RL ($2{,}321$)/evaluation($1{,}025$) splits, summarized in Table~\ref{tab:dataset_stats}.
 % summarizes the split-level statistics.
All three splits exhibit highly consistent distributions in category composition and difficulty:
\emph{Plane Geometry} accounts for about $52$--$55\%$, followed by \emph{Function} ($27$--$29\%$) and \emph{Analytic Geometry} ($18$--$20\%$), while hard problems consistently comprise about $54\%$ of each split.
% This stability ensures that training and evaluation are aligned in both geometric content and reasoning complexity.
A distinctive property of the dataset is its substantial multi-image coverage.
Across splits, $38$--$44\%$ of instances contain at least two images.
% Such problems require models to maintain semantic and geometric consistency across multiple visual contexts (e.g., referenced sub-figures or complementary views), which directly stresses the alignment dimension targeted by Faire and goes beyond single-diagram supervision.
% Such problems require models to maintain semantic and geometric consistency across multiple visual contexts (e.g., referenced sub-figures or complementary views), directly stressing the alignment conditions under which construction must function as an intermediate reasoning state.

\begin{figure}[h]
  \centering
  \includegraphics[width=0.54\linewidth]{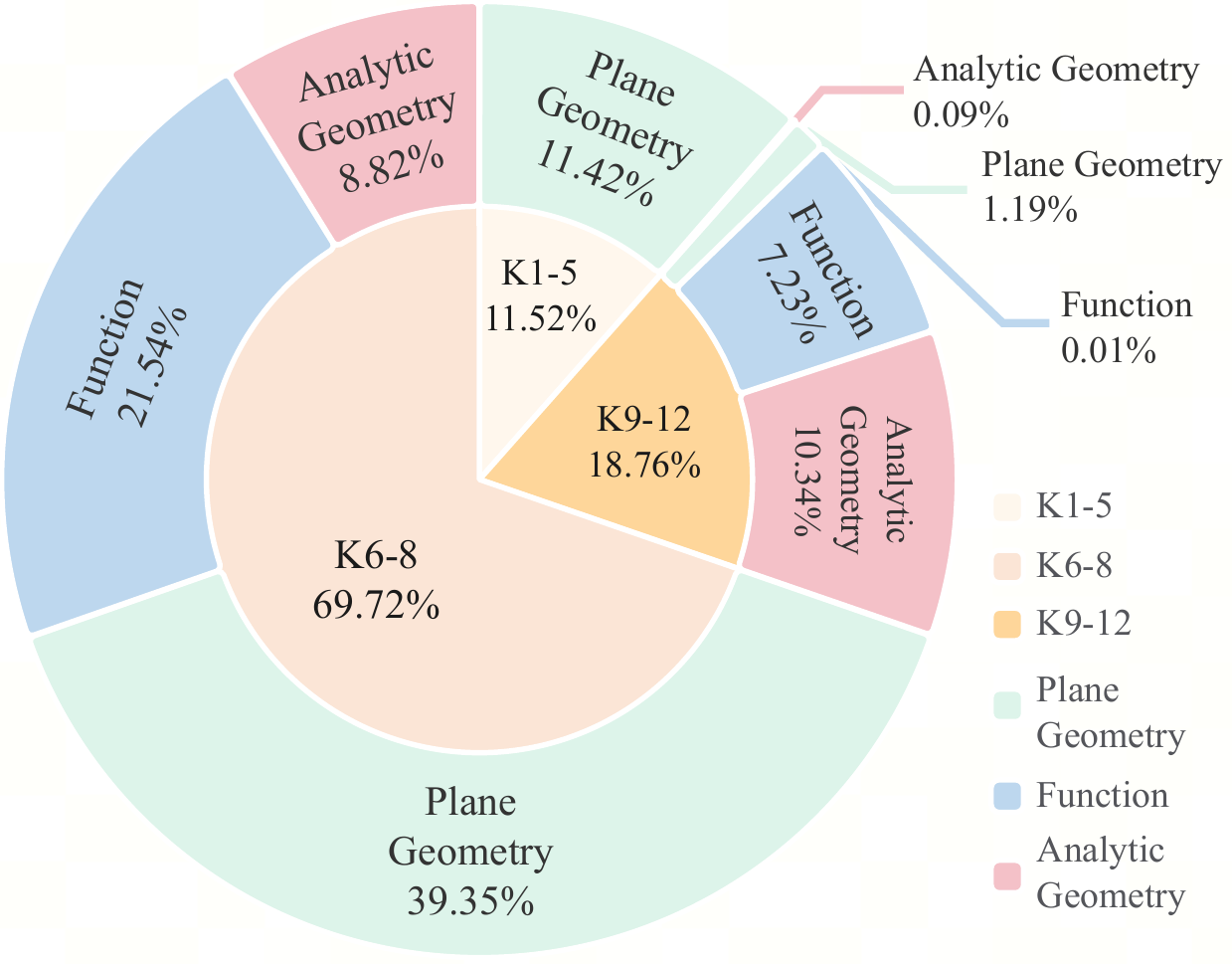}
  % \vspace{-2mm}
  \caption{Stage and category distribution.
  Inner ring shows educational stages; outer ring shows category shares within each stage.}
  \vspace{-4mm}
  \label{fig:data_analysis}
\end{figure}
\noindent\textbf{Educational stages and categories}
Figure~\ref{fig:data_analysis} shows the interaction between educational stages and geometric categories.
Middle-school content (K6--8) dominates the corpus (about $69\%$), while K9--12 contributes $17$--$20\%$ and K0--5 contributes $11$--$14\%$.
This skew reflects the diagram-required selection criterion: K6--8 is the stage where students transition from arithmetic to explicit geometric reasoning, making diagrammatic construction central rather than auxiliary.
% Within this stage, Plane Geometry remains dominant, while Function and Analytic Geometry form a substantial minority, ensuring coverage of both topological constructions and coordinate-sensitive plotting.

\noindent\textbf{Fine-grained skill coverage}
At the sub-category level, the corpus concentrates on triangle congruence and similarity ($27.8\%$), function graphs ($25.6\%$), and polygon reasoning ($14.7\%$), followed by line--coordinate geometry ($9.1\%)$ and conic sections with linear relations ($6.2\%$).
% These skills are inherently diagram-heavy and require precise constructions with faithful alignment between textual reasoning and visual evidence, reinforcing the dataset’s suitability for evaluating the trade-off between solution rigor and visual precision.
These skills are inherently diagram-heavy and require precise constructions with faithful alignment between textual reasoning and visual evidence, reinforcing the dataset’s suitability for studying when construction becomes a functional part of reasoning.

%% file: sections/050rl.tex
\section{Experimental}

\begin{table*}[t!]
  \centering
  \caption{Main results on interleaved geometry reasoning.
Acc: answer accuracy.
Verification uses tri-perspective scores: Parser ($v_{\mathrm{sem}}$), Code ($v_{\mathrm{form}}$), Judge ($v_{\mathrm{vis}}$).
Similarity metrics (BLEU/ROUGE-L/chrF; PSNR/SSIM/LPIPS) report surface matching to references.}
% Best results are bold with \textcolor{blue}{blue} background, and second-best results have \textcolor{green}{green} background.}
  \label{tab:main_results}
  \setlength{\tabcolsep}{1.5mm} % 调整列间距
  \renewcommand{\arraystretch}{1.05}
  \footnotesize
    \begin{adjustbox}{max width=\textwidth}
  % \resizebox{\linewidth}{!}{
  \begin{tabular}{l|c|ccc|ccc|cccc}
    \toprule
    \multicolumn{1}{l|}{\multirow{2}[4]{*}{Model}} &
    \multicolumn{1}{c|}{\multirow{2}[4]{*}{Acc}} &
    \multicolumn{3}{c|}{Code Similarity} &
    \multicolumn{3}{c|}{Image Similarity} &
    \multicolumn{4}{c}{Verification Scores} \\
    \cmidrule{3-12}
      & & BLEU & ROUGE-L & chrF & PSNR & SSIM & LPIPS & Parser & Code & Judge & Avg \\
    \midrule
    Gemma3-12B~\cite{team2025gemma3} & 20.97 & 7.43 & 25.35 & 33.57 & 0.83 & 3.60 & 97.56 & 1.97 & 4.13 & 0.88 & 2.33 \\
    Kimi-VL-A3B~\cite{moonshot2025kimivl} & 23.22 & 5.00 & 15.21 & 27.18 & 1.45 & 6.53 & 95.70 & 2.61 & 11.94 & 6.82 & 7.12 \\
    InternVL3.5-8B~\cite{wang2025internvl35} & 45.56 & 8.57 & 27.49 & 32.81 & 4.77 & 21.33 & 85.76 & 5.00 & 17.84 & 5.20 & 9.35 \\
    Qwen2.5-VL-7B~\cite{qwen2025qwen25vl} & 26.34 & 3.97 & 11.60 & 16.66 & 0.60 & 2.61 & 98.27 & 5.16 & 21.83 & 3.31 & 10.10 \\
    GLM-4.1V-9B~~\cite{vteam2026glm45vglm41vthinkingversatilemultimodal} & 63.41 & 5.86 & 15.70 & 23.97 & 0.67 & 2.98 & 98.05 & 7.43 & 22.39 & 7.04 & 12.29 \\
    Qwen3-VL-8B~\cite{bai2025qwen3vltechnicalreport} & 59.71 & 16.72 & 36.71 & \cellcolor[rgb]{.851,.961,.839}54.96 & 6.32 & 27.02 & 81.50 & 8.62 & 25.47 & 10.47 & 14.85 \\
    \midrule
    GPT-4o~\cite{hurst2024gpt} & 25.56 & 9.14 & 27.13 & 36.01 & 2.16 & 9.28 & 93.50 & 4.62 & 8.92 & 5.27 & 6.27 \\
    GPT-5.1~\cite{openai2025gpt51} & 56.58 & 11.85 & 32.78 & 47.08 & 7.39 & 32.50 & 76.50 & \cellcolor[rgb]{.851,.961,.839}10.34 & 25.85 & 18.24 & 18.14 \\
    Gemini-2.5-Pro~\cite{google2025gemini25} & \cellcolor[rgb]{.851,.953,.992}\textbf{78.24} & \cellcolor[rgb]{.851,.961,.839}20.20 & \cellcolor[rgb]{.851,.961,.839}38.62 & 54.70 & 4.82 & 21.67 & 85.15 & 15.90 & 22.51 & 15.45 & 19.86 \\
    GPT-5.2~\cite{openai2026gpt52} & 68.78 & 2.32 & 9.77 & 44.32 & \cellcolor[rgb]{.851,.961,.839}9.85 & \cellcolor[rgb]{.851,.961,.839}43.79 & \cellcolor[rgb]{.851,.961,.839}69.23 & 7.02 & \cellcolor[rgb]{.851,.961,.839}36.59 & \cellcolor[rgb]{.851,.961,.839}30.24 & \cellcolor[rgb]{.851,.961,.839}24.62 \\
    \midrule
    Faire (Ours) & \cellcolor[rgb]{.851,.961,.839}74.82 & \cellcolor[rgb]{.851,.953,.992}\textbf{25.06} & \cellcolor[rgb]{.851,.953,.992}\textbf{46.75} & \cellcolor[rgb]{.851,.953,.992}\textbf{55.57} & \cellcolor[rgb]{.851,.953,.992}\textbf{14.79} & \cellcolor[rgb]{.851,.953,.992}\textbf{65.10} & \cellcolor[rgb]{.851,.953,.992}\textbf{52.76} & \cellcolor[rgb]{.851,.953,.992}\textbf{37.27} & \cellcolor[rgb]{.851,.953,.992}\textbf{60.39} & \cellcolor[rgb]{.851,.953,.992}\textbf{38.44} & \cellcolor[rgb]{.851,.953,.992}\textbf{45.37} \\
    \bottomrule
  \end{tabular}
  \end{adjustbox}
  % }
  \vspace{-2mm}
\end{table*}

\subsection{Experimental Setup}
\label{sec:experimental_setup}

We evaluate Faire against a broad set of strong multimodal baselines, including both proprietary and open-weight MLLMs.
The proprietary models include GPT-4o~\cite{hurst2024gpt}, GPT-5.1~\cite{openai2025gpt51}, GPT-5.2~\cite{openai2026gpt52}, and Gemini-2.5-Pro~\cite{google2025gemini25}.
For open-weight baselines, we consider GLM-4.1V-9B~\cite{vteam2026glm45vglm41vthinkingversatilemultimodal}, Gemma3-12B~\cite{team2025gemma3}, InternVL3.5-8B~\cite{wang2025internvl35}, Kimi-VL-A3B~\cite{moonshot2025kimivl}, Qwen2.5-VL-7B~\cite{qwen2025qwen25vl}, and Qwen3-VL-8B~\cite{bai2025qwen3vltechnicalreport}.
Unless otherwise specified, all models follow the same prompting protocol and evaluation budget.

\textbf{Supervised initialization}
We initialize Faire from Qwen3-VL-8B and apply SFT with bfloat16 precision and SDPA attention.
SFT runs on a single 8-GPU node with a visual token cap of 2{,}048 and a maximum sequence length of 10{,}000 to accommodate interleaved text--code contexts.
We train for 2 epochs with learning rate $5\times10^{-6}$ and warmup ratio 0.05, and use DeepSpeed ZeRO-2 for memory optimization.

\textbf{Reinforcement learning}
Starting from the SFT checkpoint, we apply GRPO to refine interleaved trajectories under our verification signals.
We accelerate sampling with vLLM and generate 7 candidates per prompt with temperature 0.9.
% to encourage exploration.
% During training, we monitor performance on a held-out validation split and check for solutions that exploit the reward without improving geometric correctness.

\subsection{Main Results}
\label{sec:main_results}

Table~\ref{tab:main_results} reports both answer accuracy and our tri-perspective verification, which operationalizes the difference between \emph{distributional alignment} (matching the interleaving format) and \emph{functional alignment} (constructing states that subsequent deductions can verifiably rely on). Faire achieves the strongest functional alignment by a wide margin: Avg reaches 45.37, while the best proprietary baselines remain below 25 (Gemini-2.5-Pro~\cite{google2025gemini25} at 19.86; GPT-5.2~\cite{openai2026gpt52} at 24.62).
The lead is structural rather than cosmetic: Faire improves all three verification views at once, with Parser at 37.27, Code at 60.39, and Judge at 38.44. These results support the following observations.

\textbf{Accuracy leader does not imply interleaved reasoning.}
Gemini-2.5-Pro~\cite{google2025gemini25} attains the highest answer accuracy, yet its constructions fail to carry the proof burden: formal validity remains at 22.51 and the overall verification average stays at 19.86.
This discrepancy suggests that the model likely relies on textual shortcuts or internal knowledge to solve problems, treating the plotting step as a detached ritual rather than a computational tool.
In contrast, Faire pairs strong answer accuracy with substantially stronger formal grounding, nearly tripling formal validity (60.39 versus 22.51), indicating that it reasons \emph{through} verifiable constructions rather than independently of them.

\textbf{Surface mimicry vs. causal grounding.}
GPT-5.2~\cite{openai2026gpt52} often produces diagrams that visually resemble the reference, reflected in high SSIM and LPIPS scores, yet its semantic alignment collapses, with Parser dropping to 7.02.
This pattern exemplifies distributional alignment: the output matches appearance statistics while failing to instantiate the relations required by the reasoning.
Faire avoids this pitfall by prioritizing verifiable relations over surface resemblance, achieving much stronger semantic and formal consistency.
% (Parser 37.27; Code 60.39).

\textbf{RL quantifies the \textbf{Aha Moment}.}
Relative to its supervised initializer Qwen3-VL-8B~\cite{bai2025qwen3vltechnicalreport}, Faire raises the overall verification score from 14.85 to 45.37.
The largest gains appear precisely on the signals that encode functional coupling: Parser improves from 8.62 to 37.27, and Code from 25.47 to 60.39.
This quantitative leap captures the \textit{Aha Moment}: the visual generation transitions from a fragile, formatting artifact into a robust, load-bearing scaffold. The model has not just optimized a metric; it has internalized the tool, learning that correct plotting is the necessary causal antecedent to correct reasoning.
% Together, these shifts capture the Aha Moment, where construction transitions from a formatting artifact into a dependable intermediate state that actively supports subsequent deduction.

\subsection{The \textbf{Aha Moment}: Entropy Shifts under RL}
\label{sec:entropy_analysis}
\begin{figure}[ht]
    \centering
    % \vspace{-2mm}
    \includegraphics[width=0.58\linewidth]{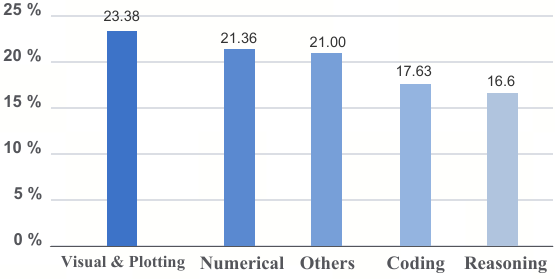}
    \caption{Distribution of top-100 entropy-increased tokens after RL.
    Tokens are grouped by semantic function.}
    \vspace{-2mm}
    \label{fig:entropy_shift}
\end{figure}

Figure~\ref{fig:entropy_shift} analyzes the top-100 tokens whose entropy increases most from SFT to RL.
Here, higher entropy does not indicate randomness.
Instead, it reflects a shift away from low-entropy, template-driven generation toward deliberate computation.
The largest entropy increase appears on \emph{visual drawing tokens} (23.38\%), showing that diagram construction is no longer executed as a fixed routine.
Under RL, drawing becomes an active decision process shaped by verifier feedback, rather than a replay of memorized patterns.
Notably, \emph{numeric} (21.36\%) and \emph{code} (17.63\%) tokens together account for nearly 40\% of the entropy shift.
% These tokens correspond to precision-critical operations.
Their entropy increase indicates that the model moves from guessing numerically plausible values to explicitly computing and validating them, a hallmark of reliable geometric reasoning.
Finally, \emph{reasoning-chain tokens} (16.60\%) also exhibit elevated entropy, suggesting that the model no longer commits early to a fixed solution path, but allows constructed diagrams to influence subsequent deductions.
We view it as the behavioral signature of the \textbf{Aha moment}, where construction becomes functionally integrated into reasoning.
% Overall, RL induces a clear cognitive reallocation:
% model capacity shifts from fluent pattern completion to visual grounding and precision logic, enabling functional alignment between drawing and deduction.

\begin{figure}[ht]
  \centering
  \includegraphics[width=0.6\linewidth]{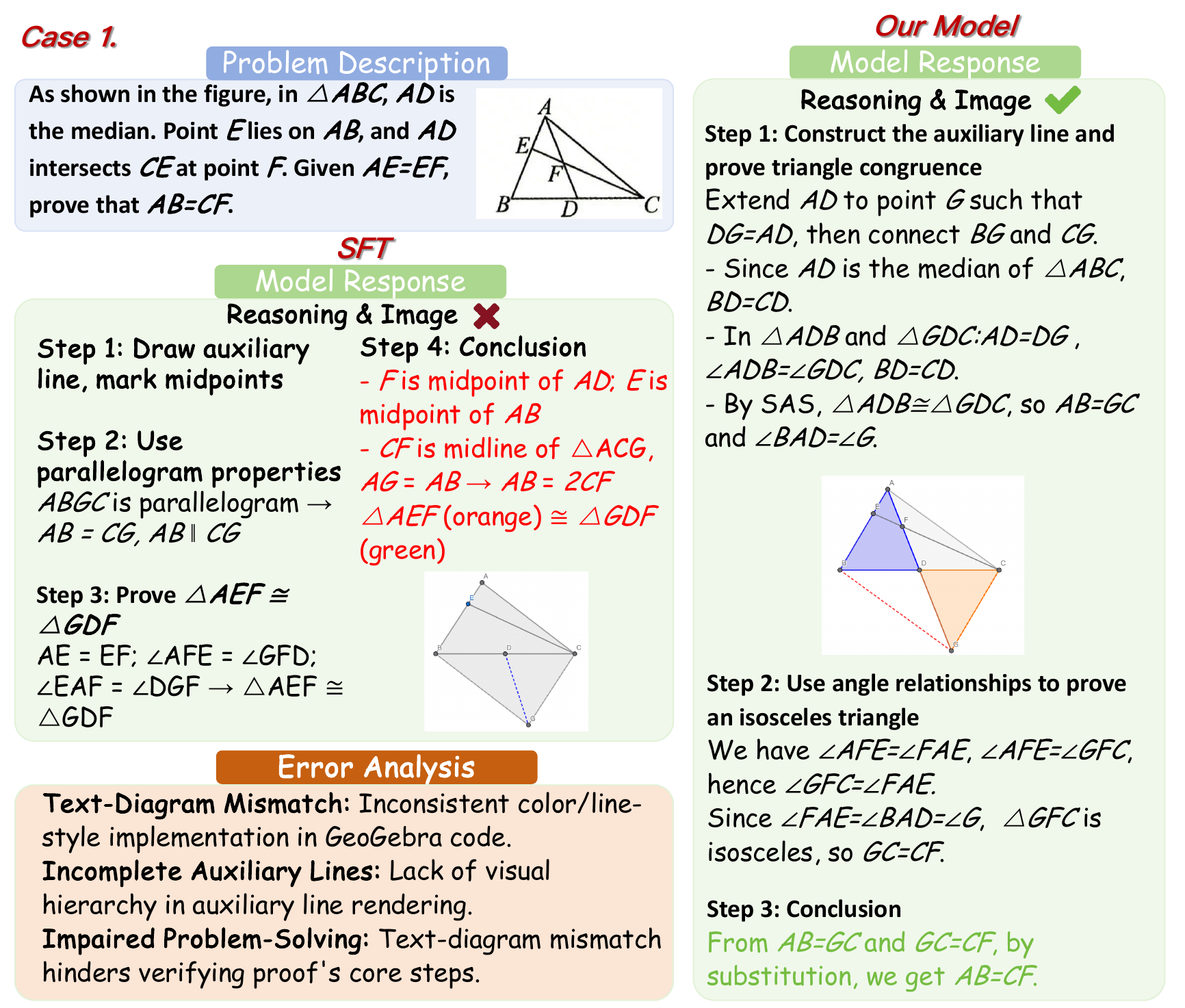}
  % \vspace{-0.5em}
  % \caption{RL vs. SFT Text–Diagram Alignment in a Geometry.}
  \caption{RL vs. SFT Text–Diagram Alignment in a Geometry problem, where RL model learns when to draw during reasoning.}
  \label{fig:case_rl_sft}
\vspace{-1.0em}
\end{figure}

\subsection{The \textbf{Aha Moment}: When Drawing Hurts No More}
\label{sec:aha_case}
\begin{figure}[ht]
  \centering
  % \vspace{-4mm}
  \includegraphics[width=0.6\linewidth]{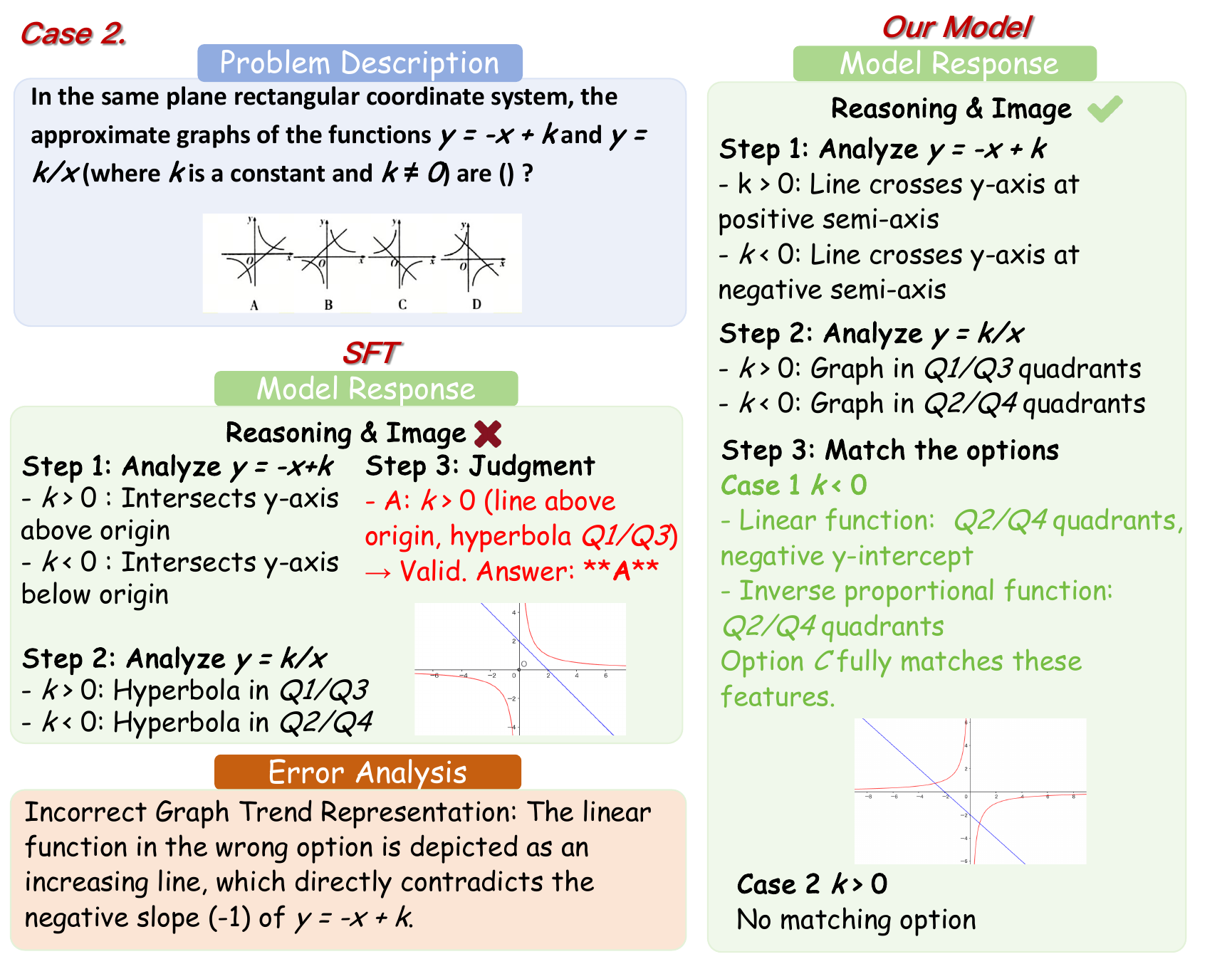}
  % \vspace{-0.5em}
  % \caption{RL vs. SFT Text–Diagram Alignment in Function Graph.}
  \caption{RL vs. SFT in Function Graph. The SFT model draws incorrect quadrant distributions with text rationales, while the RL model draws an accurate image to infer the correct answer.}
  \label{fig:case_rl_sft2}
% \vspace{-0.5em}
\end{figure}

Figure~\ref{fig:case_rl_sft}-\ref{fig:case_rl_sft2} illustrates the failure mode behind the SFT paradox and the behavioral shift induced by Faire.
With interleaved SFT, the model can mimic the \emph{format} of alternating reasoning and drawing, yet the construction is only loosely tied to the proof: the diagram highlights the claimed congruence but does not faithfully implement the intended visual hierarchy, so the “image step” cannot reliably support the “reasoning step”.
In contrast, Faire produces a construction that is operationally consistent with the proof: auxiliary points and segments are instantiated in the right dependency order, and the rendered figure functions as \emph{evidence} rather than decoration.
Under SFT, the act of drawing is a distractor that consumes context window with mismatched information. Under Faire, drawing becomes a valid working state that the subsequent deduction can rely on.
It is the Aha moment we target—drawing becomes a working state the next deduction can depend on—matching the large gains in Table~\ref{tab:main_results} and the RL-driven entropy reallocation in Figure~\ref{fig:entropy_shift}.

\subsection{The \textbf{Aha Moment}: Structure over Relations}

\begin{table}[h]
  \centering
  \caption{Results on GenExam. Math-Str: structural correctness; Math-Rel: relational correctness.}
  \label{tab:genexam}
  \small
  \setlength{\tabcolsep}{1.5mm}
  \renewcommand{\arraystretch}{1.05}

  \begin{tabularx}{0.7\columnwidth}{@{}>{\raggedright\arraybackslash\hspace{0pt}}Xcc@{}}
    \toprule
    Model & Math-Rel & Math-Str \\
    \midrule
    Emu3~\cite{wang2024emu3} & 11.3 & 0.0 \\
    Janus-Pro~\cite{chen2025janus} & 13.7 & 0.0 \\
    Qwen-Image~\cite{wu2025qwen} & 18.9 & 0.0 \\
    Imagen-4-Ultra~\cite{saharia2022photorealistic} & 35.9 & 2.6 \\
    Gemini-2.5-Image~\cite{google2025gemini25} & 43.1 & 0.7 \\
    GPT-Image-1~\cite{hurst2024gpt} &
    52.0 &
   8.0 \\
    \midrule
    \rowcolor[rgb]{.851,.953,.992}Faire &
    \textbf{52.3} &
    \textbf{9.3} \\
    \bottomrule
  \end{tabularx}
  \vspace{-3mm}
\end{table}

Table~\ref{tab:genexam} evaluates Faire on GenExam~\cite{wang2025genexam}, which separates \emph{structural correctness} (Math-Str) from \emph{relational correctness} (Math-Rel), testing whether the RL-induced behavior we target in geometric interleaved reasoning transfers beyond geometry: not just describing relations, but maintaining a coherent underlying structure.

The contrast is sharp: several strong baselines achieve high Math-Rel (e.g., GPT-Image-1~\cite{hurst2024gpt} at 52.0) while scoring 0.0 on Math-Str, indicating that relation-level plausibility can be produced without committing to a valid construction.
Faire breaks this pattern, reaching the best Math-Str (9.3) while also matching the best Math-Rel (52.3).
This supports our central claim about the \textit{Aha Moment}: RL shifts the model from imitating relational outcomes to prioritizing structure that can actually support reasoning.
Additional results on multi-step generative geometric reasoning benchmark GGBench are provided in Appendix~\ref{GGBench}.

\subsection{Ablation Study}
\label{sec:ablation}

We ablate Faire to validate two claims: (i) interleaved supervision under SFT can induce a structural failure mode (the \emph{SFT paradox}), and (ii) reinforcement learning is required to recover \emph{functional alignment} between constructed geometric states and subsequent deductions.

\textbf{Reward completeness is essential.}
Table~\ref{tab:ablation_reward} reports a leave-one-out ablation of the Faire reward.
Optimizing with an executability-only signal yields only modest improvements, indicating that RL alone is insufficient.
In contrast, removing any single verifier consistently degrades both accuracy and Draw Avg, showing that no proxy is adequate on its own and the three signals are genuinely complementary.
\begin{table}[ht]
\small
\centering
% \vspace{-2mm}
\caption{Reward ablation (leave-one-out).
Executability-only RL provides limited gains; Faire requires all verifiers to achieve functional alignment.}
\vspace{-2mm}
\label{tab:ablation_reward}
\setlength{\tabcolsep}{5.8mm}
\renewcommand{\arraystretch}{1.0}
\begin{tabular}{l|cc}
\toprule
Model & Acc & Draw Avg \\
\midrule
Faire (Full)         & \textbf{74.83} & \textbf{45.37} \\
Faire w/o Alignment  & 66.53 & 38.31 \\
Faire w/o Semantic   & 66.24 & 39.37 \\
Faire w/o Geo Assert & 65.88 & 37.15 \\
\midrule
RL (Exec-only)                & 64.29 & 35.65 \\
SFT (Interleaved)             & 62.48 & 35.23 \\
\bottomrule
\end{tabular}
\vspace{-4mm}
\end{table}

\textbf{Interleaving fails under SFT but succeeds under RL.}
Table~\ref{tab:ablation_strategy} isolates the paradox.
Interleaving under SFT reduces accuracy relative to a text-only pipeline, suggesting that the model can imitate alternation without internalizing its causal role.
After RL, the same interleaved pipeline reverses this degradation and outperforms its text-only counterpart, drawing shifts from a distraction to a usable intermediate state once the objective enforces state-to-reasoning coupling.

\begin{table}[h]
\centering
\small
\setlength{\tabcolsep}{3.2mm}
\caption{Ablation on Training Paradigms. Comparing the interplay between drawing and solving. Note that while drawing initially hurts SFT accuracy, it boosts performance post-RL.}
\label{tab:ablation_strategy}
% \resizebox{\linewidth}{!}{%
\begin{tabular}{l|cc|cc}
\toprule
\multirow{2}{*}{{Paradigms}} & \multicolumn{2}{c|}{{SFT}} & \multicolumn{2}{c}{{RL}} \\ 
 & Acc (\%) & Draw & Acc (\%) & Draw \\ \midrule
Text-only & 68.13 & - & 71.21 & - \\
% Post-hoc Drawing & 58.93 & 25.59 & - & - \\
Interleaved & 62.48 & 35.23 & \textbf{74.83} & \textbf{45.37} 
\\ \bottomrule
% \vspace{-2mm}
\end{tabular}%
% }
\end{table}

\begin{figure}[ht]
  \centering
  \vspace{-0.75em}
  \includegraphics[width=0.7\linewidth]{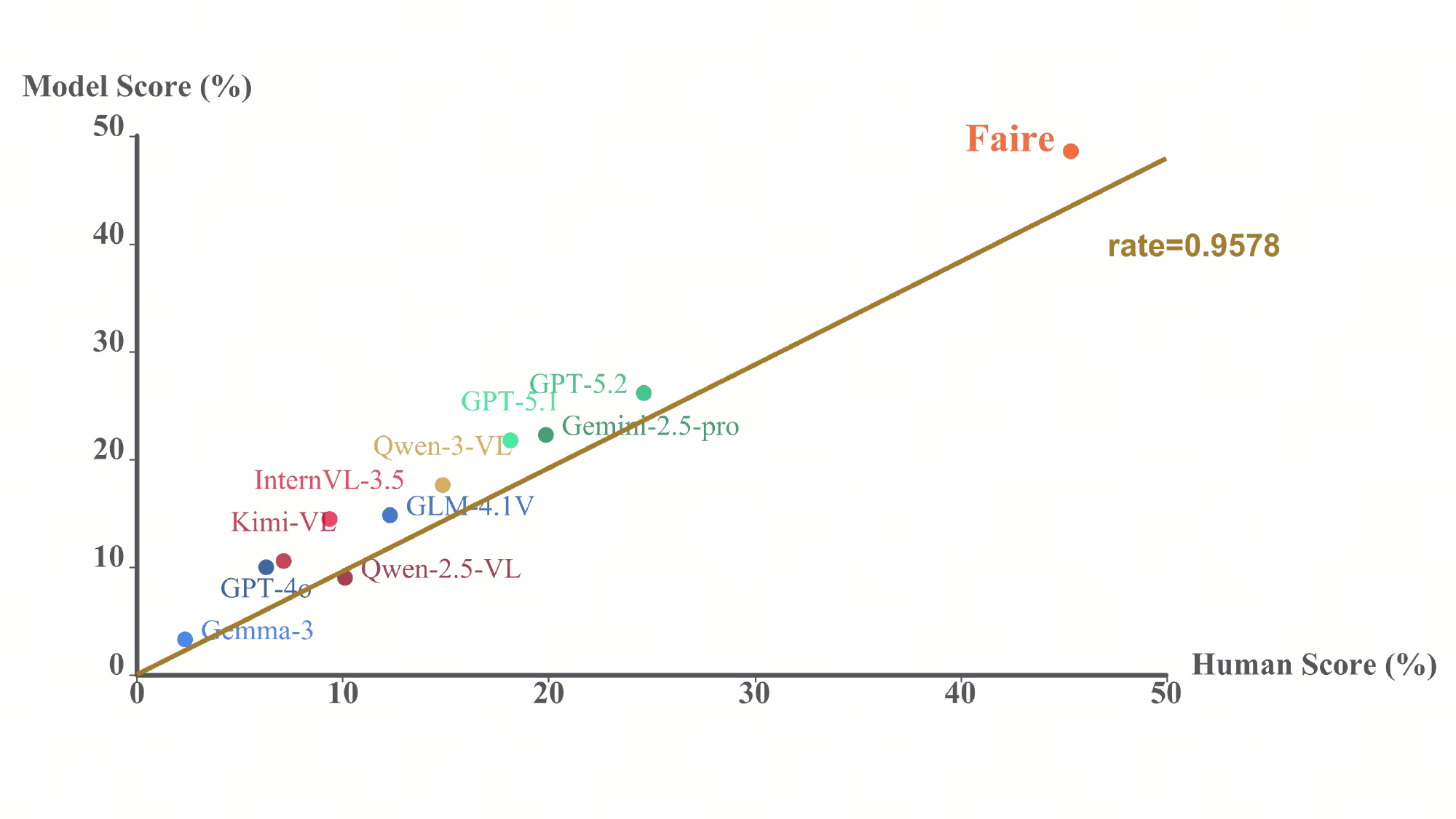}
  \vspace{-2.5em}
  \caption{Verifier score vs.\ human rating ($r=0.9578$).}
  \vspace{-0.5em}
  \label{fig:human_consistency}
\end{figure}

\subsection{Consistency with Human Judgments}
\label{sec:consistency_human}
% Figure~\ref{fig:human_consistency} shows a clear separation between prior MLLMs and Faire.
% Generalist baselines (e.g., GPT-4o~\cite{hurst2024gpt} and Gemini-2.5-Pro~\cite{google2025gemini25}) cluster in the lower-left, indicating that low verifier scores coincide with low human ratings.
% Faire is the only model in the top-right, achieving high automatic scores that are mirrored by human preference.
% The strong correlation ($r=0.9578$) provides direct evidence that our verifier score is not a brittle proxy: even after RL optimization, higher verifier scores translate into higher human ratings, suggesting limited reward hacking.
% Overall, tri-perspective verification tracks genuine diagram usability and supports Faire as a reliable training signal for interleaved geometric reasoning.
Figure~\ref{fig:human_consistency} shows a clear spatial separation.
Most existing MLLMs, including GPT-4o~\cite{hurst2024gpt} and Gemini-2.5-Pro~\cite{google2025gemini25}, cluster in the lower-left region with both low verifier scores and low human ratings.
In contrast, Faire is the only model that occupies the top-right corner, achieving simultaneously high automatic scores and high human preference.
The near-perfect correlation ($r=0.9578$) demonstrates a strong alignment between automated verification and human judgment.
Despite being optimized via reinforcement learning, verifier improvements translate monotonically into human judgment, indicating that the reward cannot be exploited by superficial artifacts.
This confirms that our tri-perspective verification captures genuine diagram usability rather than proxy signals.
Overall, Faire does not merely improve scores but establishes a distinct regime where automated verification and human evaluation are tightly aligned, validating the robustness of our reward design for interleaved geometric reasoning.

%% file: sections/100conclusion.tex
\section{Conclusion}

% We introduce a superb reasoning model named \method, which achieves excellent performance across both reasoning tasks and non-reasoning tasks. It utilizes advanced RL techniques to improve the thinking ability stably and reliably by attaining 86.7\% on AIME24, 74.0\% on AIME25 and 55.0\% on Codeforces. In the future, we plan to investigate more efficient RL recipes and explore more challenging tasks with thinking mode to push the boundary of model's intelligence. Moreover, general reward modeling with comparable accuracy as verifier would also be a compelling research direction. 

% \section{Conclusion}

% We present StructVRM, a training and verification method tailored for challenging multimodal reasoning tasks. It introduces a verifiable reward model that supports rule-based checking and model-based judgment, and leverages structured data augmentation to improve robustness during reinforcement learning. StructVRM demonstrates strong performance across multiple reasoning benchmarks, and provides a scalable basis for fine-grained supervision in complex, real-world settings.
Geometric problem solving often requires \emph{interleaved reasoning}, where diagram construction and logical deduction must support each other step by step.
We surface a counter-intuitive \emph{SFT paradox}: supervised fine-tuning on interleaved traces can reduce solving accuracy, because it fits the \emph{distribution} of alternation while leaving the construction--reasoning dependency under-optimized.

We address this gap with Faire, which enforces \emph{functional alignment} via reinforcement learning with gated rewards and post-generation verification from complementary visual, semantic, and formal verifiers.
Across benchmarks and categories, Faire turns construction into a dependable intermediate state for the next deduction, capturing the Aha Moment that interleaving is meant to enable.
Importantly, the verifier-driven scores track human judgments closely, supporting that the learned gains reflect better grounding rather than reward hacking.
We hope this work encourages verifiable interleaved reasoning beyond geometry, and motivates stronger assertion synthesis with executable tools.